\documentclass[acmsmall]{acmart}
\usepackage{algorithm,algpseudocode} 
\usepackage{caption}
\usepackage{subcaption}
\usepackage{graphicx}
\usepackage{amsmath}

\AtBeginDocument{%
  \providecommand\BibTeX{{%
    \normalfont B\kern-0.5em{\scshape i\kern-0.25em b}\kern-0.8em\TeX}}}

\begin{document}

%%
%% The "title" command has an optional parameter,
%% allowing the author to define a "short title" to be used in page headers.
\title[Optimal Bidding Strategy using NFQ]{Neural Fitted Q Iteration based Optimal Bidding Strategy in Real Time Reactive Power Market}

%%
%% The "author" command and its associated commands are used to define
%% the authors and their affiliations.
%% Of note is the shared affiliation of the first two authors, and the
%% "authornote" and "authornotemark" commands
%% used to denote shared contribution to the research.
\author{Jahnvi Patel}
%\authornote{Both authors contributed equally to this research.}
\email{jahnvi.1210@gmail.com}
\affiliation{%
  \institution{Department of Computer Science and Engineering \& Robert Bosch Centre for Data Science and AI, IIT Madras}
}
%\orcid{1234-5678-9012}
\author{Devika Jay}
%\authornotemark[1]
\email{devikajay@gmail.com}
\affiliation{%
  \institution{Department of Electrical Engineering, IIT Madras}
}

\author{Balaraman Ravindran}
\affiliation{%
  \institution{Department of Computer Science and Engineering \& Robert Bosch Centre for Data Science and AI, IIT Madras}
}
\email{ravi@cse.iitm.ac.in}

\author{K. Shanti Swarup}
%\authornotemark[1]
\email{swarup@ee.iitm.ac.in}
\affiliation{%
  \institution{Department of Electrical Engineering, IIT Madras}
}

%%
%% By default, the full list of authors will be used in the page
%% headers. Often, this list is too long, and will overlap
%% other information printed in the page headers. This command allows
%% the author to define a more concise list
%% of authors' names for this purpose.
%\renewcommand{\shortauthors}{Trovato and Tobin, et al.}

%%
%% The abstract is a short summary of the work to be presented in the
%% article.
\begin{abstract}
In real-time electricity markets, the objective of generation companies (GENCOs) while bidding is to maximise their profit. The strategies for learning optimal bidding have been formulated through game theoretical approaches and stochastic optimisation problems. In game theoretical approaches, payoffs of rivals are assumed to be known, which is unrealistic. In stochastic optimisation methods, a suitable distribution function for rivals' bids is assumed. The uncertainty in the network is handled by assuming the process to be a Markov Decision Process with known state transition probabilities. Similar studies in reactive power markets have not been reported so far because the network voltage operating conditions have an increased impact on reactive power markets than on active power markets. Contrary to active power markets, the bids of rivals are not directly related to fuel costs in reactive power markets. Hence, the assumption of a suitable probability distribution function is unrealistic, making the strategies adopted in active power markets unsuitable for learning optimal bids in reactive power market mechanisms. Therefore, a bidding strategy is to be learnt from market observations and experience in imperfect oligopolistic competition based markets. In this paper, a pioneer work on learning optimal bidding strategies from observation and experience in a three-stage reactive power market is reported. Such bidding strategy studies on reactive power markets have not been studied extensively in the literature. Also, the learning method proposed in this paper does not assume any probability distribution to model the uncertainties that affect one's bidding strategy. For learning optimal bidding strategies, a variant of Neural Fitted Q Iteration (NFQ-TP) with prioritized experience replay and target network is proposed. The total reactive power requirement is estimated by the learning agent through a Long Short Term Memory (LSTM) network to suitably define the state space for the learning agent. The learning technique is tested on three-stage reactive power mechanism of IEEE 30-bus Power System test bed under different scenarios and NFQ network configurations. The simulation results ensure that the technique is suitable for learning optimal bidding strategies from their own experiences and market observations.
\end{abstract}

%%
%% The code below is generated by the tool at http://dl.acm.org/ccs.cfm.
%% Please copy and paste the code instead of the example below.
%%
\begin{CCSXML}
<ccs2012>
   <concept>
       <concept_id>10010405.10010432.10010439</concept_id>
       <concept_desc>Applied computing~Engineering</concept_desc>
       <concept_significance>300</concept_significance>
       </concept>
   <concept>
       <concept_id>10010147.10010257.10010258.10010261</concept_id>
       <concept_desc>Computing methodologies~Reinforcement learning</concept_desc>
       <concept_significance>500</concept_significance>
       </concept>
   <concept>
       <concept_id>10010147.10010341.10010342</concept_id>
       <concept_desc>Computing methodologies~Model development and analysis</concept_desc>
       <concept_significance>100</concept_significance>
       </concept>
 </ccs2012>
\end{CCSXML}

\ccsdesc[300]{Applied computing~Engineering}
\ccsdesc[500]{Computing methodologies~ Reinforcement learning}
\ccsdesc[100]{Computing methodologies~Model development and analysis}
%%
%% Keywords. The author(s) should pick words that accurately describe
%% the work being presented. Separate the keywords with commas.
\keywords{Reactive Power Market, Optimal Bidding Strategy, Deep Reinforcement Learning, Neural Fitted Q Iteration, Prioritized Experience Replay, LSTM}

%%
%% This command processes the author and affiliation and title
%% information and builds the first part of the formatted document.
\maketitle

\section{Introduction}
With the advent of advanced information technology and integration of small-scale renewable energy sources to power supply systems, significant changes in power supply grid operations and planning are being implemented worldwide. Deregulation of vertically-integrated-utility based grid operations is considered as a major revolution in power sector. This has increased the participation of private companies in maintaining the grid operation levels within permissible limits. The grid operator functions as an Independent System Operator (ISO) that procures energy from private producers and schedules generation and demand in real-time. This forms an electricity market.

Ancillary services, like network voltage support through reactive power procurement, spinning reserve maintenance, etc., are also being achieved through market mechanisms. Among these services, reactive power procurement has been regarded as one of the most important ancillary service market \cite{us1996promoting}. However, implementing a real-time pricing mechanism for reactive power has been challenging due to the localised nature of reactive power. Reactive power supply is to be provided locally to support reactive power demand and maintain system-wide bus voltage magnitudes within permissible limits. However, this local requirement of reactive power results in the exercise of market power by participants. Under critical operating conditions, participants may submit high bids resulting in highly volatile price signals in the market. Hence mechanisms that preserve incentive compatibility and individual rationality is essential for reactive power markets \cite{9179006}. Such incentive-compatible markets provide limited information to market participants, and hence learning optimal bidding strategies is a challenge.

Most of the work so far on optimal bidding strategies for reactive power markets focuses on game theoretic approaches. Optimal bidding strategy for reactive power market was studied under conjectured supply function models with multi-leader follower game setting \cite{oligopoly}. Estimation of electricity suppliers' cost function in day-ahead real power market using inverse optimization method based on historical bidding data is discussed in \cite{past_bids}. In \cite{mul_leader}, the behaviour of generation companies participating in reactive power market is simulated as a multi-leader follower game.  In \cite{colombian}, Q-learning is used to strategize bidding in market models based on DC-OPF. Deep Q Networks (DQNs) are used to find the optimal bidding strategy for generators in single-sided energy markets as in \cite{learn}. 

Bidding strategies discussed in the literature so far considered a simple pricing mechanism where, in most of the cases, uniform price signals were issued to generation companies without considering the location features and real-time reactive power requirements. However, these bidding strategies fail when reactive power pricing mechanism issues nodal prices, i.e., prices that vary with the location of the generation companies. This makes price signals highly correlated with preceding operating conditions like active power demand from loads, network topology changes, and voltage magnitude profiles. Also, the actual cost function in the case of reactive power generation of rival generation companies cannot be estimated directly from fuel costs. The cost of reactive power generation constitutes only a small percentage of fuel cost, and fuel costs are directly related to active power generation. 

\begin{figure}[h]
    \centering
    \includegraphics[width=0.45\textwidth]{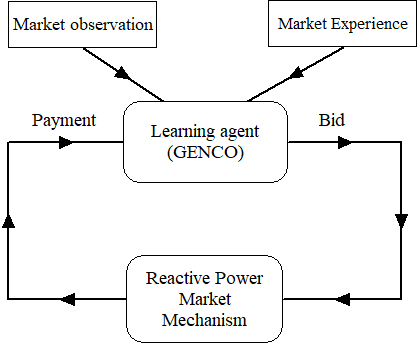}
    \caption{Learning optimal bidding strategies in reactive power market}
    \label{fig:learning}
\end{figure}

Hence, learning agents, i.e., generation companies, are exposed to an environment that provides limited access to information regarding market pricing mechanism. Imperfect information setup in the market and dependency of the current state of the market on the preceding sequence of states make learning bidding strategies a challenge. Assumption of suitable probability distribution function to model system uncertainties and bids of rivals is unrealistic in reactive power markets due to the inherent features of the market discussed above. Learning bidding strategies in an incentive-compatible three-stage reactive power market mechanism \cite{9179006} has not been reported so far. This paper presents a pioneer work in this regard by proposing a deep reinforcement learning technique for a single learning agent in an environment (three-stage reactive power market mechanism) constituting competing generation companies that do not share information. The proposed learning technique is simplified in Fig.\ref{fig:learning}.
The learning agent (GENCOs) acts on the environment (reactive power market mechanism) by submitting its optimal bids. The feedback that it receives from the environment is the payment, which is the reward signal. The additional inputs that the learning agent needs to learn an optimal bidding strategy are market observations and experience. This defines the framework of the learning technique that is detailed in this paper.

The main contributions of this work are summarised below:
\begin{enumerate}
    \item Design of an optimal bidding strategy suitable for incentive-compatible reactive power markets without assuming unrealistic probability distribution function for rivals' bids and system uncertainties. The imperfect oligopoly competition among participants adds further complexity to the market environment. Neural Fitted Q Iteration with Target network and Prioritized Experience Replay (NFQ-TP) learning algorithm is proposed to achieve optimal bidding strategy from market observations and experience for a reactive power producer such that the profit of generator is maximized. The proposed NFQ-TP increases learning stability using target networks and improves sampling efficiency using prioritized experience replay. 
    \item The market environment that can be formulated as a higher-order Markov Decision Process is converted to a first-order Markov Decision process, by suitable definition of state space in which total reactive power requirement in the system is predicted using an LSTM network.
    \item Present an approach using Long Short Term Memory networks (LSTM) to estimate current state requirement of reactive power based on total quantity generated by GENCOs in past episodes.
\end{enumerate}

The rest of the paper is organised as follows: A review of the application of deep learning techniques in the electricity market is presented in Section 2. Section 3 provides a brief background on the reactive power market model in which the bidding strategies are to be learned by the market participant. The learning environment of the market participant (agent) is proposed in Section 4. Section 5 presents the proposed Neural Fitted Q Iteration with Target Network and Prioritized Experience Replay algorithm for learning optimal bidding strategies. Experimental results and observations are discussed in Section 6. In Section 7, brief conclusions are drawn and future work discussed.

\section{Related work}
Participants like generation companies, customers, etc., have been learning bidding strategies since the implementation of electricity markets based on competitive bidding. The participants aim to utilise the varying load scenarios to earn profit by submitting optimal bids. Several works related to optimal bidding strategies were formulated as an optimisation problem that was solved using heuristic methods \cite{wen2001genetic,badri2013security}.  To consider the uncertainty of a rival's behaviour while determining bidding strategies, minimisation of normal probability distribution function of the rival's bid is generally adopted. This minimisation problem has been solved using gravitational search algorithm \cite{singh2019optimal}. Determining monthly bidding strategies was formulated as a bi-level problem in \cite{8723337}. Based on conditional value at risk, a stochastic bi‐level optimization was proposed \cite{rayati2019optimal} for coordinated wind power and gas turbine units in the real‐time market.

Day-ahead bidding problem can also be considered as a Markov decision process (MDP). In \cite{wozabal2020optimal}  a variant of the stochastic dual dynamic programming algorithm was used to solve the MDP. With advancements in artificial intelligence,
reinforcement learning has found significant application in electricity market modeling \cite{nanduri2007reinforcement,rahimiyan2010adaptive}. Deep Deterministic Policy Gradient with prioritized experience replay was proposed for complex energy markets \cite{ye2019deep}. Asynchronous advantage actor-critic (A3C) method was proposed in \cite{cao2020bidding} to determine optimal bidding strategies for wind farms in short term electricity markets.

The aforementioned methodologies that determine the optimal bidding strategies are suitable for participants in energy markets. There has been no extensive work on the formulation of an optimal bidding strategy for participants in reactive power markets. This is mainly because of the complexity associated with reactive power market. These complexities arise with difficulty in determining the actual requirement of reactive power in the system. In the case of energy markets, the quantity to be produced by the participant can be directly determined from load forecasting techniques due to active power load pattern features. Whereas in reactive power markets, the requirement of reactive power in the system is determined by reactive power loading pattern and other features like active power loading, network topology, system operating conditions like voltage, frequency, etc. and the location of the participant. Also, estimating the cost function of rivals through probability distribution function or supply function methods is not easy in reactive power markets. This is because the cost function of market participants are not directly related to fuel, contrary to energy markets. 

Hence, the general assumption of modelling system uncertainties and bids of rivals with a suitable probability distribution function and then, defining optimal bidding strategy as a solution to a stochastic optimisation problem is unrealistic in reactive power market mechanisms. Assuming reactive power market as a Markov Decision Process (MDP) is also not valid as state variables in reactive power market are highly correlated with not only the current state but also the past sequences of states. Also, incentive compatibility and limited information setup in the market to mitigate the exercising of market participants make learning optimal bidding strategies a challenge. Thus, these issues lead to the requirement of formulating a new framework for learning optimal bidding strategies for participants in reactive power market based on market observation and experience. 

\section{Reactive power market mechanism}

Design of reactive power markets involves mechanism definition that will effectively coordinate the  interaction between several generating companies (GENCOs) and Independent System Operator (ISO). In this work, a three-stage reactive power market mechanism is considered \cite{9179006} in which GENCOs submit their bids in the format as given below:
\begin{itemize}
	\item Operation cost (\$/MVAr)
	\item Lost opportunity cost (LOC, \$/MVAr$^{2}$)
\end{itemize}

\begin{figure}[h]
    \centering
    \includegraphics[width=0.75\textwidth]{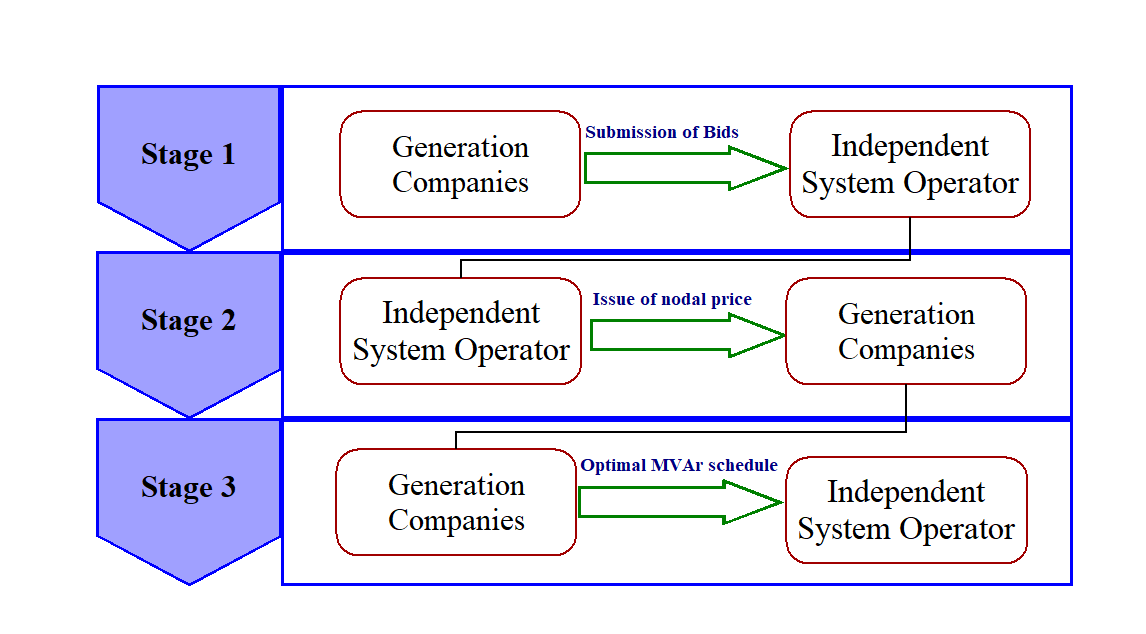}
    \caption{Three stage reactive power market model}
    \label{fig:model}
\end{figure}

Based on the bids received, ISO issues a price signal to each participant. The participants then respond with their optimal generation schedule. The stages of the market mechanism are summarised below and depicted in Fig. \ref{fig:model}.

\begin{itemize}
    \item At the start of every hour, reactive power generators, i.e., market participants, submit their price bids to the ISO for the next hour. The two dimensional bids have the form $(b_{1i}, b_{2i})$: $b_{1i}$ is the operation cost (or the linear term in the cost function of participant) and $b_{2i}$ is the lost opportunity cost (or the quadratic term in the cost function of participant).
    \item ISO responds with individual price signals to each of the GENCOs such that total payment is minimized while meeting the demand and other criterion.
    \item GENCOs respond by submitting the quantities they would generate at the next time step. The last two steps are repeated till all the market conditions have been satisfied.
\end{itemize}

As mentioned, each GENCO receives a separate price signal, also known as nodal pricing, and the generation scheduling is done based on the cost function of each GENCO. The cost function of each GENCO is private information and is not revealed. Also, estimating the cost function of rivals from fuel costs, as in energy markets, is not suitable in the reactive power market. This is because the cost of reactive power is not directly related to fuel costs. These market features pose challenges in learning due to partial observability. Also, the GENCOs have limited knowledge on the network topology  and system operating conditions that have resulted in different price signals at different nodes of the network. Reactive power requirement in the system depends on voltage profile in the system, in addition to network topology and apparent power demand from loads. Hence, the conditional probability distribution of future states depends on a sequence of preceding states. Thus, the agents have a partially observable environment with imperfect competition based market mechanism to learn their optimal bidding strategy.
% Reactive power requirement in the system depends on voltage profile in the system in addition to network topology and apparent power demand from loads. Hence the conditional probability distribution of future states depend on a sequence of preceding states. This makes the general assumption of considering thee environment for learning optimal bidding strategies as a Markov Decision Process (MDP) invalid. Hence a learning methodology for bidders that compete in an partially observable market mechanism considering higher order MDP is required so as to earn profit.

\section{Reactive power market environment model for learning}
\label{sec:env}
As described in earlier sections, learning of optimal bidding strategy in reactive power market becomes practical only when the learning process considers the following features for market environment:
\begin{enumerate}
    \item Higher order Markov Decision Process (MDP)
    \item Imperfect information setup
    \item Imperfect competition
\end{enumerate}

Higher-order MDP formulation and imperfect information in the market are to be handled so that the market environment translates to a reinforcement learning setup. The feature of imperfect competition is discussed in detail in IEEE 30-bus system simulation studies. 

Modelling of reactive power market environment to suit reinforcement learning setup is presented in this section. This section describes key definitions crucial to the discussion of methodologies presented in the paper and formulates a first-order MDP for the optimal bidding problem. 
\subsection{Handling higher order Markov Decision Process}
\label{sec:mdp}
 In a reinforcement learning setting, an MDP is characterized by a set of states $S$, a set of actions $A$, transition probability function $p(s_{t+1}|s_t,a_t)$ satisfying Markov property that describes the probability of the learning agent being in state $s_{t+1}$ on taking action $a_t$ in state $s_t$ and an immediate reward function $S \times A \rightarrow R$. At any time step, the goal of the learning agent is to choose an action $a_t$ according to a policy $\pi^*(s_t)$ that maximizes the expected return $G_t = r_{t+1} + \gamma r_{t+2} + \gamma^2 r_{t+3} +\ldots$ where $\gamma$ is the discount factor that determines the trade-off between long term and short term rewards. This is done by following a policy that maximizes the value of action value function $Q(s_t, a_t) = E[G_t | s_t, a_t, \pi]$ for the given state $s_t$.

\begin{figure}[h]
    \centering
    \includegraphics[trim=40 40 40 84,clip, width=0.55\linewidth]{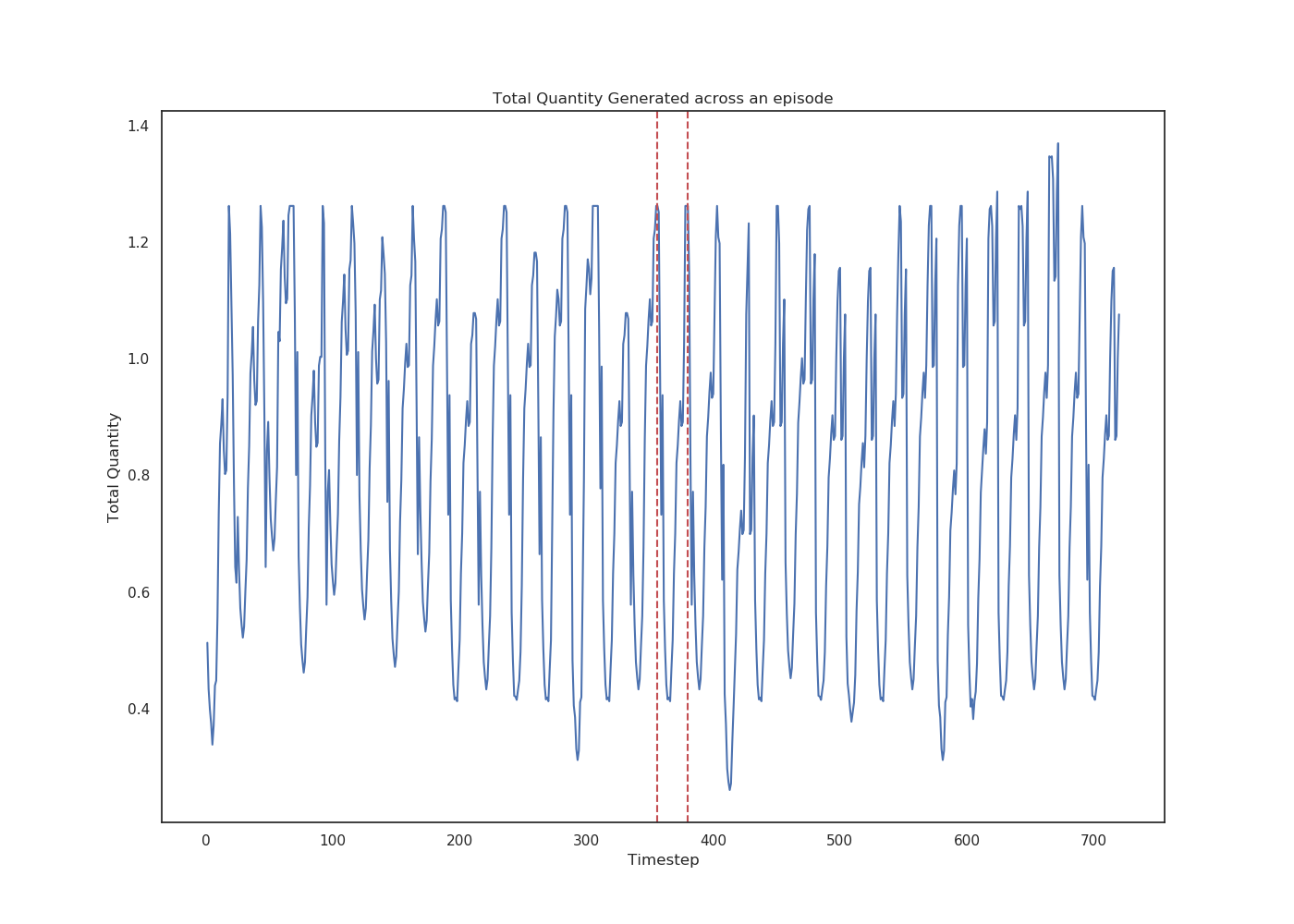}
    \caption{Total Quantity generated by all GENCOs across an episode}
    \label{fig:qty_time_curve}
\end{figure}

Translating the above to the proposed market setting, the GENCO for which optimal bidding strategy is to be learnt from market observations constitutes the learning agent. Stages 2 and 3 of the three-stage market mechanism (Figure~\ref{fig:model}) can be encapsulated into an optimization model such that when any GENCO submits its price bids to the model (stage 1), it receives the price signal and generation schedule for the next hour. This optimization model is the environment in our RL setting. Next, we analyze the market properties in order to fabricate the state features accurately. Under the assumption that reactive power requirement at any given time step remains almost the same irrespective of GENCOs behaviour, we fix the bids of all GENCOs to an arbitrary value and calculate the total quantity generated by all GENCOs in the market at each time step. Figure \ref{fig:qty_time_curve} shows how total quantity generated by all reactive power producers (a measure of reactive power requirement) varies with time across an episode. The dotted red lines are separated temporally by 24 hours. We observe a heavy correlation in loads at time-step $t-48, t-24, t-2, t-1$ to the current load. Therefore, it is possible to evaluate the price signals and generation schedule in a similar market situation from our experience and use them to predict the optimal behaviour in the current market scenario. 

Utilising signals from preceding time steps to describe the next state vector means that the probability of transitioning into the next state (time $t$) depends not only on the current state (time $t-1$), but also on the sequence of states preceding it, specifically $t-2, t-24, t-48$ in this case. This violates the first-order Markov assumption that \textit{future is independent of the past given the present} which presents a major hurdle in formulating the problem as an MDP. However, such seemingly non-Markov cases such as ours when the next step depends on a bounded sequence of steps from past, i.e. $p(s_{t+1} | s_t, a_t, s_{t-1}, a_{t-1}, \ldots) = p(s_{t+1} | s_t, a_t, \ldots, s_{t-48}, a_{t-48})$, then such a higher-order Markov process can be translated to a first-order Markov problem by augmenting the state vector and embedding the past (48 hours) as an additional feature. Doing so makes the next step depend only on the current state and the action taken, thereby making it possible to formulate an MDP for the problem. 
\subsection{Handling imperfect information in the market}
\label{lstm}
As mentioned in earlier sections, the reactive power requirement in the system depends on not only the reactive power loads in the network but also the network topology. This poses an imperfect information setup in the learning environment. Load prediction techniques and probability based assumptions will not be a direct input to determine the total quantity of reactive power generation required in the system. Thus, the reactive power requirement at each time step is to be estimated from the total quantity of reactive power generated in the network in previous time steps to see how the reactive power requirement at current time-step correlates to the market experience we have gained so far. 

For this, consider the time series data of total quantity generated by all GENCOs in IEEE 30-bus system from an earlier episode (month) as the raw training input. From the market analysis in Section \ref{sec:mdp}, it is known that this data is periodic in nature, which makes it suitable for using Recurrent Neural Networks (RNNs) which is a class of neural networks that allow information to persist and hence, make it easier to learn sequential data. An unrolled RNN can be thought of as multiple sequential copies of an Artificial Neural Network. However, this poses an issue when it comes to long term memory. RNNs consider only recent information while forgetting temporally separated events that could link to current output. This is where Long Short Term Memory (LSTM) Networks, a special class of Recurrent Neural Networks, come in to learn long term dependencies. 

LSTMs accumulate an internal state that is constantly updated using the inputs, a hidden state that can be used for computing output, forget gate that determines what information to throw away from the cell state and together, the hidden state, input, and cell state can be used to compute the next hidden state. Many-to-one LSTM models take a series of recent observations as input and generate a single output corresponding to prediction for the next time step. Hence, the learning agent could use LSTMs to forecast load for the next hour as a sequence prediction.

The raw reactive power requirement (total quantity) signal is transformed into the training set for LSTM by splitting the input into sequence of length 24 corresponding to the reactive power requirement observations over the past one day. The target is defined as the total quantity for the next hour. Inputs from the training set are then fed into an LSTM network with $n$ units followed by a dense layer. The network is trained by back-propagating the mean squared loss between predicted reactive power requirement and true reactive power requirement. The pre-trained LSTM network can be used to estimate the total quantity required in the current state for state representation of the learning agent. The LSTM reactive power requirement prediction algorithm is outlined in Algorithm \ref{alg:lstm_pred}.

\begin{algorithm}
	\caption{Reactive Power Requirement Prediction using LSTM}
	\label{alg:lstm_pred}
	\begin{algorithmic}[1]
	\State Let $S$ $\leftarrow$ Total quantity time series data for a month
	\State Initialize $X, y$ $\leftarrow$ Split $S$ into sequence of length 24 and reactive power requirement for next hour as corresponding output 
	\State Define model $M \leftarrow$ [LSTM(n), Dense]
		\For {$iteration=1,2,\ldots$}
			\State Sample a batch $B$ from the training set
			\State Use $M$ to generate the predicted output values $\hat{y_i}$ for $i=1..B$
			\State Train the network using Adam optimizer on $\sum_i(y_i - \hat{y_i})^2$ for $i=1..B$ and update the LSTM weights
		\EndFor
	\end{algorithmic} 
\end{algorithm}

\subsection{Formulating the Reinforcement Learning problem}
Having performed suitable translation of higher order MDP to first order MDP, and handling imperfect information setup in the market environment, the features of the Reinforcement Learning based agent is described as an MDP $(S, A, P, R, \gamma)$.

\textbf{State Space:} The state features are designed in such a way that the agent can use them to determine the consequences of its actions while also integrating current market conditions like the current reactive power requirement estimate. A description of the state features can be found below:

\begin{enumerate}
    \item \textbf{Previous bids:} This considers the past experience of the learning agent in the market to learn the optimal bidding strategy. We consider the ratio of bids sent to ISO to actual cost function of the learning agent $i$ : $<a1_{i,t'}, a2_{i,t'}> \forall t' \in \{t-48, t-24, t-2, t-1\}$ timestamps. 

\item \textbf{Reward signals:} The rewards received at $\{t-48, t-24, t-2, t-1\}$  timesteps act as feedback signal to the agent and will help the learner to judge the optimality of bids. A lower reward will indicate that the learner bid too high resulting in lesser payment. This is because the generation scheduled through stages 2 and 3 of the market mechanism was less due to high bid.

\item \textbf{Total Quantity Estimate:} We use the previous quantity signals to estimate the total generation at the current time-step, which is correlated to the reactive power requirement at that timestep. LSTM network discussed in Section \ref{lstm} could be utilized to predict this periodic quantity from time-series data.
\end{enumerate}

\textbf{Action Space:} The bids that we send to the optimization module constitute the action for that state. For ease of analysis, instead of considering bids as action signals directly, we consider the action to be bid magnification defined as the ratio of bids sent to ISO to actual production cost. Thus, for $i^{th}$ learner at timestep $t$ with actual operation cost $c1_{i}$ and lost opportunity cost $c2_{i}$, the action space is (a1$_{i,t}$, a2$_{i,t}$) defined as:
\begin{equation}
    <a1_{i,t}, a2_{i,t}> = <\frac{b1_{i,t}}{c1_{i}}, \frac{b2_{i,t}}{c2_{i}}>,
\end{equation}
where $b1_{i,t}$ and $b2_{i,t}$ represent the bid submitted to ISO as the claimed operation and lost opportunity cost respectively. An advantage of using the ratios over actual bid values is that the actions can now remain same across all GENCOs leading to easier analysis.  Because our action space is low-dimensional, it is possible to discretize it at the required granularity and can be effectively considered as discrete space. In order to bound the action space, we restrict the bid magnification coefficients to $[1,5]$ range and discretize the action space using step size of 0.5. This results in a total of 81 discrete actions : $(1,1), (1, 1.5), \ldots, (2.5,2.5), \ldots, (5, 4.5), (5,5)$. 

\textbf{Reward Function:} The reward is measured in terms of profit made by reactive power producer as compared to bidding its true cost. Let $c1_i$ and $c2_i$ represent the cost coefficients for agent $i$, $qg$ be the quantity generated by the producer and $bg_i$ be the base generation. Base generation of a generation company is the amount of reactive power required for the auxillary services within the plant and also for shipment of base active power produced. Thus, we express the profit of GENCO as follows:
\begin{equation}
    p_{i,t} = price_{i,t} \times qg_{i,t} - c1_i \times (qg_{i,t}-bg_i) - c2_i \times (qg_{i,t}-bg_i)^2 
\end{equation} 
The reward signal is computed as $r_{i,t} = p_{i,t} -p^b_{i,t}$, where $p^b_{i,t}$ is computed using price and generation returned from the environment upon sending the bids $(b1_{i,t}, b2_{i,t}) = (c1_i, c2_i)$ to the ISO optimization model. By using a reward measure relative to baseline, the learner would have an incentive to learn actions that perform better than bidding their true costs and also be penalized by a negative reward when they receive payment less than baseline. 

\section{Neural Fitted Q Iteration for learning from market observation and experience}
In this section, a variant of Neural Fitted Q Iteration (NFQ), i.e., NFQ with Target Network and Prioritized Experience Replay (NFQ-TP), is proposed for the single learning agent to handle the imperfect competition and incomplete information of the market environment. Brief background of classical Q-learning and NFQ is first discussed before presenting the proposed learning workflow and NFQ-TP algorithm.
\subsection{Classical Q-Learning}
In order to solve the reinforcement learning problem, the agent should learn the expected return for each state-action pair. Q-learning \cite{watkins1992q} is a model-free, off-policy, temporal difference based learning algorithm that uses Bellman Optimality Equation [\ref{eq:bellman}] to recursively update the Q-values until convergence is reached.
\begin{equation}
    Q(s_t, a_t) = r_{t+1} + \gamma \max_{a'} Q(s_{t+1}, a')
\label{eq:bellman}
\end{equation}
Q-learning updates estimates based on learned estimates using bootstrapping. Let learning rate be $\alpha$, then the classical Q-learning update rule after $k$ iterations is given by [\ref{eq:ql_update}].
\begin{equation}
    Q_{k+1}(s_t, a_t) = (1-\alpha) Q_k(s_t, a_t) + \alpha (r_{t+1} + \gamma \max_{a'} Q_k(s_{t+1}, a'))
\label{eq:ql_update}
\end{equation}
The term $\delta_t = r_{t+1} + \gamma \max_{a'} Q_k(s_{t+1}, a') - Q_k(s_t, a_t)$ is known as the TD-error and it is a measure of the error in our current estimate of Q-values. In spite of having theoretical convergence properties, this kind of tabular approach is suitable only for small and finite state spaces. This is because the algorithm becomes computation and memory intensive as the number of states increases. 

However, the 13-dimensional state space defined for the learning agent is continuous, and discretization of it leads to an exponential increase in state space size. Thus, classical Q-learning technique is unsuitable for the learning agent in an imperfect, oligopolistic reactive power market. Hence, deep reinforcement techniques like Neural Fitted Q Iteration are required for learning optimal bidding strategies.

\subsection{Neural Fitted Q Iteration}
Classical Q-learning algorithm suffers from the curse of dimensionality. Also, updating the Q-function based on only one point can lead to slower and noisy convergence.
The challenges posed by the classical Q-learning technique can be overcome by using Fitted Q Iteration (FQI) \cite{ernst2005tree}, a batch mode reinforcement learning algorithm suitable for continuous state spaces and is an extension of the traditional Q-learning approach. Basis functions are needed to map the input states into the derived features space, which will approximate optimal Q values. We can then use any supervised learning technique, such as regression or SVM, to fit the training set and generalize this information to any unseen state. At each step, the learning agent has access to the experience buffer from which it samples a mini-batch as its training set. The advantage of doing this is that the same experience tuple can be used for learning multiple times for learning while also breaking temporal correlations in sequentially generated data, thereby satisfying the i.i.d (independently identically distributed) assumption. The training set is of the form $(<s, a>, Q)$ with the state-action pairs as input to the algorithm and the target being defined as the action-value $Q(s,a)$ for the state-action pair. 

Neural Fitted Q Iteration (NFQ) \cite{riedmiller2005neural} is a data-efficient, deep learning based algorithm belonging to the FQI family. It builds on the FQI algorithm by using the global generalization effects of a neural network as a regressor to approximate the Q values using offline data. The principle of NFQ is the same: using a single point to update the weights can create unintended changes in the weights for other state-action pairs, whereas a batch based approach would stabilise learning. NFQ uses RPROP as an optimizer to minimize oscillations in back-propagation. The advantage of using resilient back-propagation \cite{riedmiller1993direct} is that it adapts the step size for each weight independently based on the sign of the double derivative of the loss function. Summarizing, if $\theta$ are the weights of the NFQ network, then corresponding to the input state-action pair $<s, a>$, the output of the network would be the action value $Q(s, a; \theta)$. Since NFQ has been shown to perform well in control problems with continuous state spaces, it is suitable for learning the Q-value function for the optimal strategy problem. 

\subsection{Proposed NFQ-TP based learning algorithm}
This section presents a variation of NFQ algorithm, i.e., NFQ with target network and prioritized experience replay (\textit{NFQ-TP}) is proposed. The advantage is that the proposed algorithm leads to faster convergence for the bidding environment. 

\textbf{Prioritized experience replay:}
The original NFQ algorithm works on a pre-generated offline training data sample of size $N$ using random trajectories. However, populating the buffer sufficiently with all patterns of state-action pairs is difficult while using random actions because of the high-dimensional state space and large number of actions. Hence, in NFQ-TP, the experiences are iteratively added to the buffer at each step of training.
Initially, the buffer is populated with $N_I$ initial episodes generated using random trajectories. During the training phase, to balance exploitation with exploration, the agent adds an experience tuple by stepping in the environment according to an $\epsilon$-greedy exploration strategy. This means that with $(1-\epsilon)$ probability, the agent chooses an action that maximizes action-value $Q(s, a; \theta)$ predicted by \textit{NFQ-TP} for the given state and with $\epsilon$ probability chooses an action randomly. The exploration factor $\epsilon$ is initially set to a large value to encourage taking rare actions and decayed as the algorithm approaches convergence.  

The sampling efficiency from experience replay can be further improved compared to uniform sampling by picking the important transitions with high TD-error more frequently using a Prioritized Experience Replay \cite{schaul2015prioritized}. Higher magnitude of
TD-error means the network does not estimate the Q-value for that experience tuple accurately. Hence, we want to pick such samples more frequently. Each experience 
in the buffer is stored along with the priority $p_i = |\delta| + \epsilon_P$  where $|\delta| = $ is the magnitude of TD-error and $\epsilon_P$ is a constant added to ensure that each experience is picked with a non zero probability. The sampling probability of a transition $P(i)$ is defined as:
\begin{equation}
    P(i) = \frac{p_i^\beta}{\sum_{j=1}^N p_j^\beta}
\end{equation}
where $\beta$ controls the amount of prioritization with 0 corresponding to uniform sampling and 1 corresponding to complete priority based sampling. \\
\textbf{Updating target network:}
NFQ uses current Q estimates to predict the target Q value for loss computation, and since these weights are constantly updated at each step in the learning process, the target values are prone to oscillations. To stabilize learning further, a target network is used with the parameters from the target network being updated slowly using parameters from the local network. Therefore, for each training step, once the local weights $\theta$ are updated, the weights for the target network are updated as:
\begin{equation}
    \theta^T \leftarrow (1-\tau) \theta^T + \tau \theta
\end{equation}
where  $\theta^T$ and $\theta$ are parameters for target and local network respectively and $\tau << 1$ decides the rate at which target network changes. The NFQ target would now be computed using the target network weights as:
\begin{equation}
    y_t = r_t + \gamma \max_{a'} Q(s_{t+1}, a';\theta^T)
\end{equation}
The TD-error $\delta_t$ for any transition can be defined as:
\begin{equation}
    \delta_t = r_{t+1} + \gamma \max_{a'} Q(s_{t+1}, a';\theta^T) - Q(s_t, a_t;\theta)
\end{equation}

In each learning iteration, the prioritized experience replay is replenished with new experiences by stepping in the environment in an $\epsilon$-greedy manner by using the Q-value estimate from the target network to find the maximizing action. The weights of the local network are updated by sampling a mini-batch, using the local network to estimate Q-values for each experience tuple in the batch, and performing back-propagation on mean squared TD loss with TD target computed using the target network. The weights of the target network are updated slowly based on the updates on the local network. This process is repeated until convergence is reached.
The pseudo-code for the NFQ-TP implementation is described in Algorithm \ref{alg:nfq_pseudo}.

\begin{algorithm}
	\caption{Neural Fitted Q Iteration (NFQ-TP variation)} 
		\label{alg:nfq_pseudo}

	\begin{algorithmic}[1]
	\State Initialize prioritized replay memory D to capacity N by taking random steps in environment
	\State Initialize local network with random weights $\theta$
	\State Initialize target network with weights $\theta^T \leftarrow \theta$
	\State Initialize PER parameters $\beta$ and initial priority $p_m$  
		\For {$iteration=1,2,\ldots$}
			\State Step in environment using policy $\pi_{\theta^T}$ for $K$ time steps with $\epsilon$ noise
			\State Add $K$ experience tuples of the form $(s_t,a_t,s_{t+1}, r_{t+1})$ to D and initialize priorities to  $p_m$
			\State Sample a mini-batch of size $M$ from PER
			\State Compute $y_t = r_t + \gamma \max_{a'} Q(s_{t+1}, a';\theta^T)$ for each tuple in batch
			\State Train the network using Rprop on $(y_t-Q(s_t, a_t;\theta))^2$ and update the local network $\theta$
			\State Update priorities of the sampled experiences in D
			\State Update target network $\theta^T \leftarrow (1-\tau) \theta^T + \tau \theta$
		\EndFor
	\end{algorithmic} 
\end{algorithm}

\begin{figure}[h]
    \centering
    \includegraphics[width=0.8\linewidth]{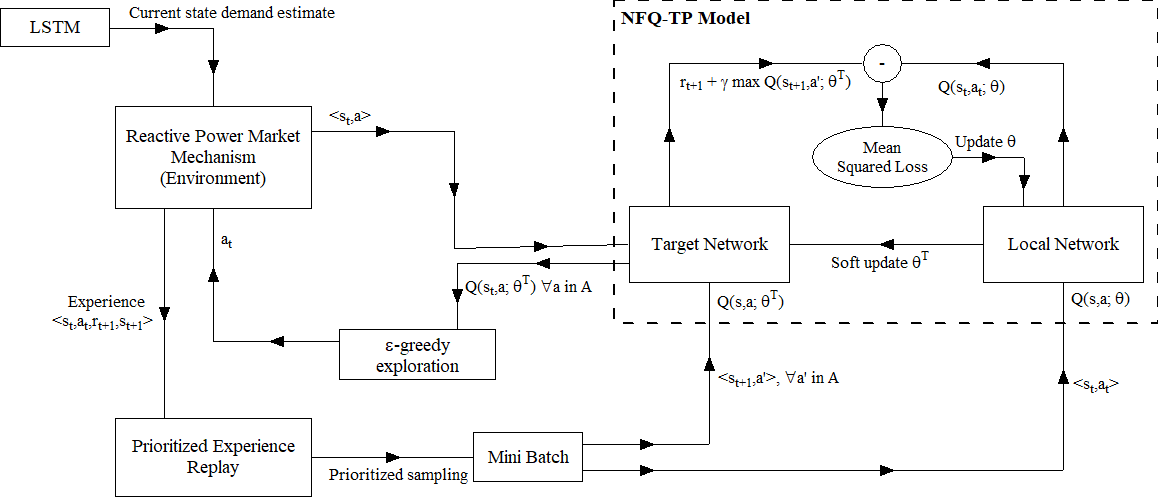}
    \caption{Optimal Bidding Strategy Training Flow}
    \label{fig:algo_flow}
\end{figure}

The learning workflow for the overall algorithm is described in Figure \ref{fig:algo_flow}. 

\section{Experiments}
\subsection{Testing framework}
\begin{figure}[h]
    \centering
    \includegraphics[width=0.45\linewidth]{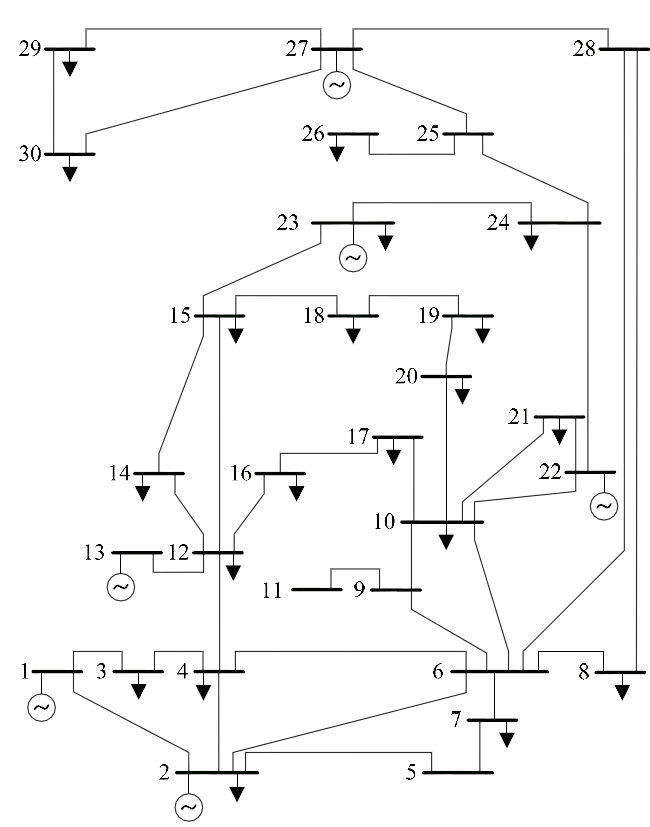}
    \caption{IEEE 30-bus system}
    \label{30bus}
\end{figure}

The three stage reactive power model was tested in IEEE 30-bus system  \cite{zimmerman2010matpower} as shown in Fig. \ref{30bus}
There are generator buses at nodes 1, 2, 13, 22, 23, 27. The peak load in the system is 189.5MW, 107.2MVAr. Each generator bus is considered as a generation company. Thus, there are 6 GENCOs in the network, competing with each other in the reactive power market. The operation cost $c1$ and lost opportunity cost $c2$ of the GENCOs are described in Table \ref{tab:cost_coeffs}. In the consecutive stages, a single leader multi-follower game mechanism results in issuing price signals to GENCOs by ISO so that the GENCOs respond with their optimal generation schedule. The base generation values for these GENCOs are tabulated in \ref{tab:base_generation}.  Each episode is a maximum of 720 ($24 \times 30$) steps long, where each step corresponds to an hourly bid in the 30-day window.

\begin{table}[htbp]
    \caption{Actual cost coefficients for all 6 GENCOs}
    \centering
    \begin{tabular}{ccl}
    \toprule
        GENCO & c1 (\$/100MVAr) & c2(\$ / (100MVAr)$^{2}$)\\
    \midrule
        1 & 0.73& 0.30  \\
        2 & 0.68 & 0.39 \\
        3 &0.75 &  0.43\\
        4 & 0.60& 0.5 \\
        5 &0.75 &0.9\\
        6 &0.73 &0.38\\
    \bottomrule
    \end{tabular}
    \label{tab:cost_coeffs}
\end{table}

\begin{table}[htbp]
    \caption{Base MVAr generation in IEEE 30-bus system}
    \centering
    \begin{tabular}{cl}
    \toprule
        GENCO & Base MVAr\\
    \midrule 
        1& 7.5 \\
        2 &  3 \\
        3& 3.125\\
        4& 2.435\\
        5& 2\\
        6& 2.235\\
    \bottomrule
    \end{tabular}
    \label{tab:base_generation}
\end{table}

The following studies have been performed to understand the performance of the proposed NFQ-TP based learning algorithm for optimal bidding in reactive power market of IEEE 30-bus system. 
\begin{enumerate}
    \item Single agent NFQ-TP based learning considering two cases of bidding strategies for rivals
    \begin{itemize}
        \item Imperfect competition where GENCOs bid higher than the actual cost (Bidding Strategy-B1)
        \item Perfect competition where GENCOs bid their actual cost (Bidding Strategy-B2)
    \end{itemize}
    The goal is to analyze the robustness of the learning approach in multiple market scenarios against differing rival bidding strategies.
    \item Comparison of two network configurations of NFQ-TP network:
    \begin{itemize}
        \item NFQ-1: Input- State features concatenated with discrete action. Output- Single Q-value for state-action pair.
        \item NFQ-2: Input- State features. Output-81 Q-values corresponding to each action.
    \end{itemize}
    Comparison of the two variants gives us insights on how to represent the Q-values effectively while permitting as much generalization as possible without impacting the representation strength of the network.
    \item Comparison of learning benefits for low cost, intermediate cost and high cost agents from proposed NFQ-TP based learning algorithm in order to analyze how the learning approach provides an improvement over the naive strategy for different learners.
\end{enumerate}

\textbf{LSTM Reactive Power Requirement Prediction.} Before we can generate the experience tuples for training \textit{NFQ-TP} network, the LSTM network must be pre-trained to be able to predict the current state reactive power requirement. We use the algorithm specified in Section \ref{lstm} and feed the input as the total quantity generation data for a 30-day long episode. We employed a network with 100 LSTM units followed by a dense layer whose output is a single value corresponding to the reactive power requirement prediction for the next time step. Since the algorithm splits the time series data into segments lasting 24 hours, the input to the LSTM network would be of size 24, corresponding to the total quantities produced by all GENCOs for each hour over the last day. 
\subsection{Defining bidding strategies for GENCOs other than NFQ-TP learning agent}
The bidding strategies adopted by the agents other than learner are described below:
\begin{enumerate}
\item {\textbf{Imperfect competition (Bidding Strategy B-1).} Optimal bidding strategies in reactive power market for producers have not been studied extensively. Hence, for GENCOs other than the learning agent, we modify the strategies adopted for bidding in active power markets. The optimal bidding strategy in active power markets is formulated as a stochastic optimisation problem by assuming suitable probability distribution functions to model uncertainties in the network and bids of rivals in the market. Such a basic optimisation problem formulation for active power market discussed in \cite{mathur2017optimal} is rewritten in the context of reactive power market as given below:
\begin{equation}
    \max_{b1_{i,t},b2_{i,t}} \quad E[\pi_{i,t}] 
\end{equation}
\textit{such that}
\begin{equation}
    \pi_{i,t} = (b1_{i,t}+b2_{i,t}\cdot qg_{i,t}) \cdot qg_{i,t} - (c1_{i,t}+c2_{i,t}\cdot qg_{i,t}) \cdot qg_{i,t}
\end{equation}
\begin{equation}
  qg_{i,t}^{min} \leq qg_{i,t} \leq qg_{i,t}^{max}  
\end{equation}
\begin{equation}
  R_{i}^{min} \leq (b1_{i,t}+b2_{i,t}\cdot qg_{i,t}) \leq R_{i}^{max}  
\end{equation}
Where R$_{i}^{min}$ and R$_{i}^{max}$ are the limits set to the bidding function which can be assumed based on the capacity of each GENCO. The optimisation problem can be solved through heuristic methods like Monte Carlo methods, Genetic Algorithm etc provided assumptions on probability distribution function for system uncertainties and rival's bids can be validated. However, as detailed in previous sections, assumption of such probability distribution function for reactive power markets is unrealistic. Hence, in this work, we consider the results reported in \cite{mathur2017optimal} for IEEE 30-bus system under peak load conditions. The reported results are applied to reactive power markets by considering demand fluctuations in the bids. For a more competitive environment, the bids in reactive power market are made higher than optimal bids in active power market. This is a fair baseline as the chances of exercising market power in reactive power is higher and the uncertainty in network operating conditions is handled by utilising demand fluctuation signals instead of assuming a  probability distribution function.

Thus, for this bidding strategy, we assume that all the agents except learner send bids equal to two times actual operation cost and five times lost opportunity cost multiplied with a step-wise variable $d_t$ between $[0,1]$ corresponding to the demand fluctuations.
\begin{equation}
    \begin{aligned}
        a1_{i,t} &= 2 \times d_t \\
        a2_{i,t} &= 5 \times d_t
    \end{aligned}
\end{equation}
Note that we use actual demand and some random noise to generate the multiplier to ensure that other GENCOs do not have perfect information.}

\item \textbf{Perfect competition (Bidding Strategy B-2).} All the agents except learner send bids equal to the actual operation and lost opportunity cost.
\end{enumerate}

\subsection{Defining variants of NFQ Network}
Two variants of NFQ network are considered to study the performance of the proposed NFQ-TP based learning. Both the variants use ReLU activation, mean squared TD-error for updating the parameters and RPROP as an optimizer. 
\begin{enumerate}
    \item \textbf{NFQ-1} takes the state features along with the discrete action as input and outputs a single Q-value for the state action pair. Since the state-space is 13-dimensional and action space is 1-dimensional, the input to NFQ-1 network would be a size 14 vector. Figure \ref{fig:nfq_var1} summarizes NFQ-1 implementation. 
\begin{figure}[h]
    \centering
    \includegraphics[width=0.5\linewidth]{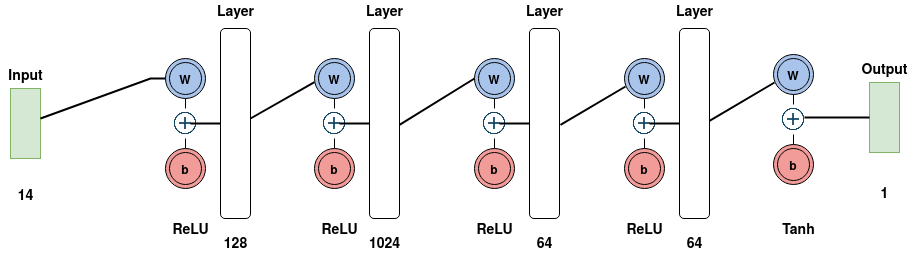}
    \caption{Visualization of NFQ network used in Variant 1}
    \label{fig:nfq_var1}
\end{figure}
\item \textbf{NFQ-2} takes the state features as input and outputs 81 Q-values corresponding to each of the actions. Input to NFQ-2 would be a size 13 vector. Figure \ref{fig:nfq_var2} depicts NFQ-2 network implementation respectively. 
\begin{figure}[h]
    \centering
    \includegraphics[width=0.5\linewidth]{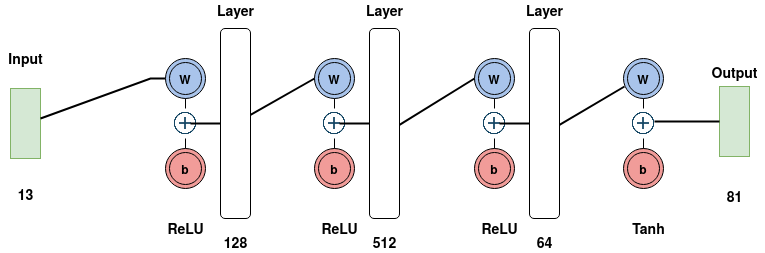}
    \caption{Visualization of NFQ network used in Variant 2}
    \label{fig:nfq_var2}
\end{figure}
\end{enumerate}
Adam optimizer is used for learning NFQ network weights with a learning rate of 0.001. A decay rate of 0.01 is used for decaying the exploration noise and a soft update rate of $\tau = 10^{-3}$ is used for updating the target network. A mini-batch of size 64 is sampled from the Prioritized Experience Replay and the prioritization factor $\beta$ is set to 0.7. The buffer capacity is set to $N= 10^5$ with number of initial experiences $N_I = 10^4$. A more thorough discussion on choosing the hyper-parameters can be found in Section \ref{sec:hyp_tune}.

\subsection{Simulation results, observation and inference}
The parameters affecting the experiment are studied in Appendix \ref{sec:hyp_tune}, that presents a brief summary on the technique used to determine the optimal set of hyper-parameters. The experiments are carried out in accordance with the results obtained from the aforementioned analysis.

\subsubsection{\textbf{LSTM Reactive Power Requirement Prediction.}} In order to evaluate the model performance, we define the baseline model that predicts current quantity as an average of total quantity generated at $t-1$ and $t-24$, i.e. $\hat{Q_t} = \frac{Q_{t-1} + Q_{t-24}}{2}$. On evaluation, the mean squared error between predicted and actual demand for baseline was found to be 0.019 whereas in case of LSTM model, the error was 0.01. Thus, it is observed that LSTM outperforms baseline and hence, using an LSTM leads to more accurate reactive power requirement prediction.

\subsubsection{\textbf{Case 1: Rivals adopt bidding strategy B-1}}
\begin{figure}[h]
    \centering
    \includegraphics[width=0.65\linewidth]{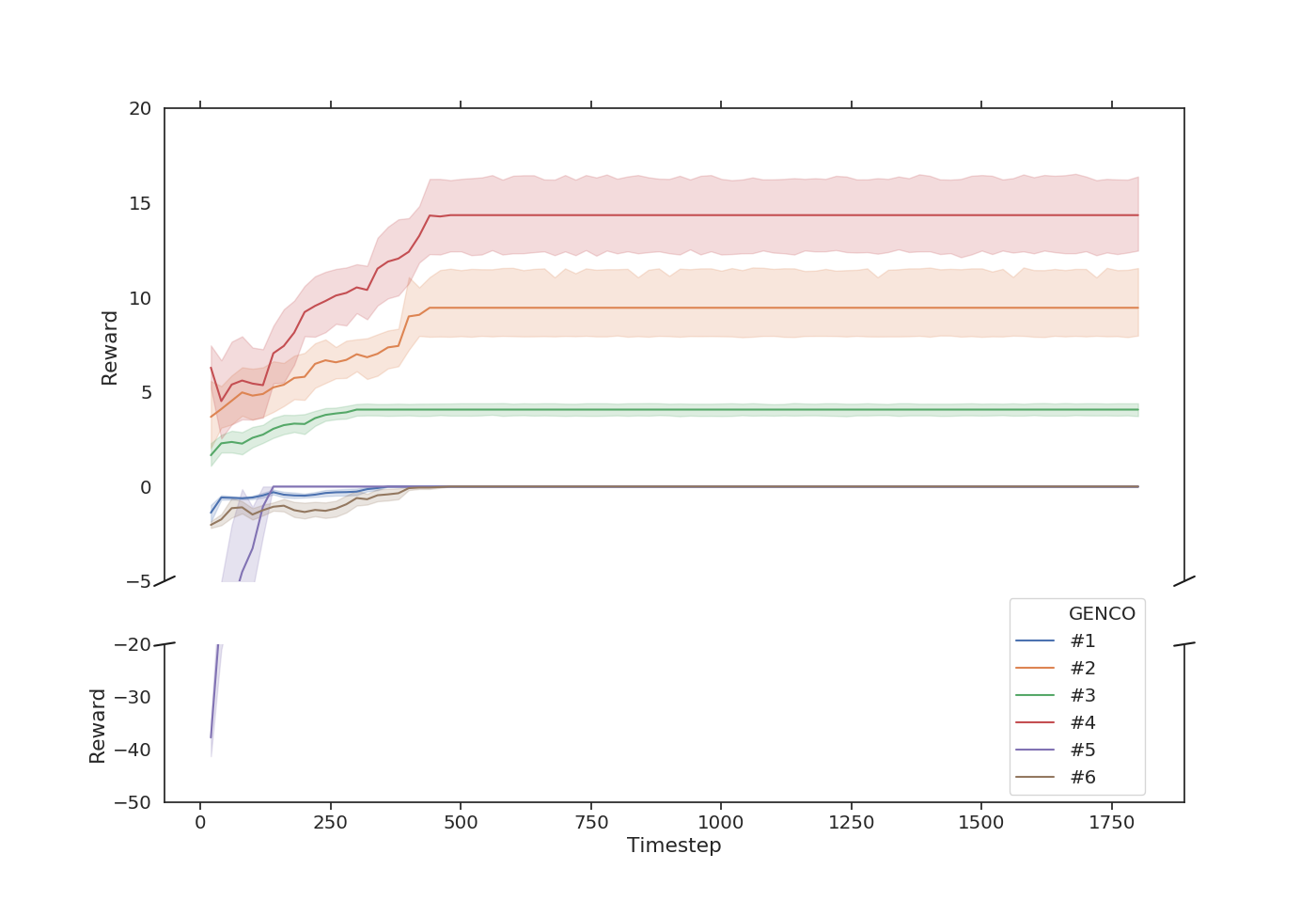}
    \caption{Learnt episodic rewards averaged across 10 random seeds with each of the GENCOs as a learner using NFQ-1 model and B-1 rival bidding strategy}
    \label{fig:rew_nfq1_b1}
\end{figure}

\begin{figure}[h]
    \centering
    \includegraphics[width=0.6\linewidth]{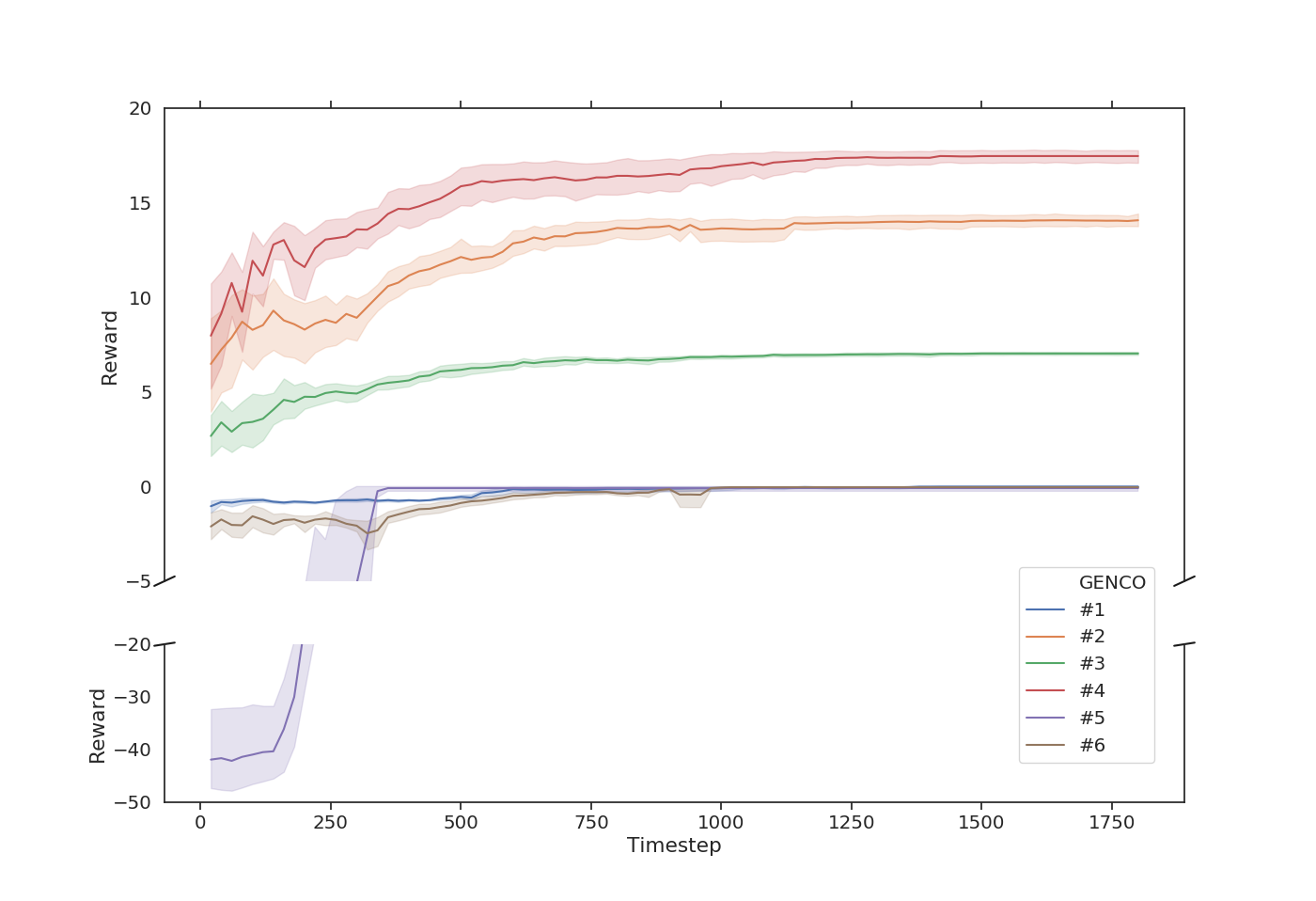}
    \caption{Learnt episodic rewards averaged across 10 random seeds with each of the GENCOs as a learner using NFQ-2 model and B-1 rival bidding strategy}
    \label{fig:rew_nfq2_b1}
\end{figure}

We consider each of the GENCOs as a learner and the bidding strategies of remaining 5 competing GENCOs are determined by B-1. This section presents the improvements achieved by each of the implementations of \textit{NFQ-TP} over the baseline true cost bidding.\\\\
\textbf{Performance of NFQ-1 variant}\\
Figure \ref{fig:rew_nfq1_b1} and Table \ref{tab:rew_nfq1_b1} show the average and standard deviation of episodic rewards across 10 random runs of the \textit{NFQ-TP} algorithm with each of the GENCOs as a learning agent and using NFQ-1 network implementation and B-1 rival bidding. The table values have been computed considering only the convergence values of episodic returns for each of the runs.
\begin{table}[htbp]
    \caption{Mean $\mu$ and standard deviation $\sigma$ over 10 random runs using NFQ-1 model and B-1 rival bidding strategy}
    \centering
    \begin{tabular}{ccccccc}
    \toprule
        \# & 1 & 2 & 3 & 4 & 5 & 6\\
    \midrule
        $\mu$ & 0.0 & 9.45 & 4.06 & 14.34 & 0.0 & 0.002\\
        $\sigma$ & 0.005 & 2.83 & 0.52 & 3.22 & 0.0 & 0.003\\
    \bottomrule
    \end{tabular}
    \label{tab:rew_nfq1_b1}
\end{table}\\
\textbf{Performance of NFQ-2 variant}\\
Figure \ref{fig:rew_nfq2_b1} and Table \ref{tab:rew_nfq2_b1} show the average and standard deviation of episodic rewards across 10 random runs of the \textit{NFQ-TP} algorithm with each of the GENCOs as a learning agent and using NFQ-2 network implementation and B-1 rival bidding.

\begin{table}[htbp]
    \caption{Mean $\mu$ and standard deviation $\sigma$ over 10 random runs using NFQ-2 model and B-1 rival bidding strategy}
    \centering
    \begin{tabular}{ccccccc}
    \toprule
        \# & 1 & 2 & 3 & 4 & 5 & 6\\
    \midrule
        $\mu$ & 0.0 & 14.08 & 7.03 & 17.47 & 0.0 & 0.0 \\
        $\sigma$ & 0.002 & 0.52 & 0.13 & 0.53 & 0.25 & 0.05\\
    \bottomrule
    \end{tabular}
    \label{tab:rew_nfq2_b1}
\end{table}

\subsubsection{\textbf{Case 2: Rivals adopt bidding strategy B-2}}
\begin{figure}[h]
    \centering
    \includegraphics[width=0.55\linewidth]{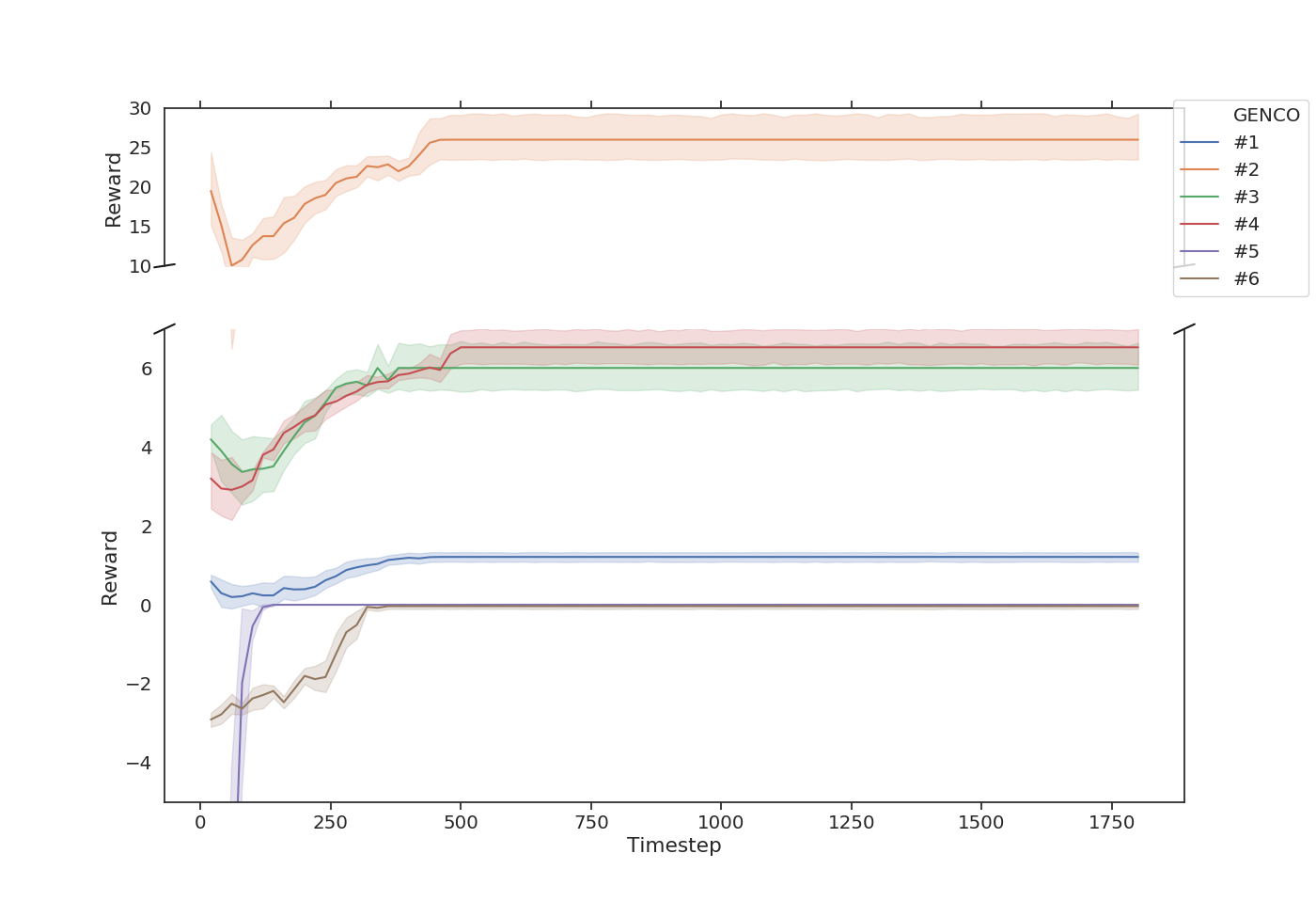}
    \caption{Learnt episodic rewards averaged across 10 random seeds with each of the GENCOs as a learner using NFQ-1 model and B-2 rival bidding strategy}
    \label{fig:rew_nfq1_b2}
\end{figure}

\begin{figure}[h]
    \centering
    \includegraphics[width=0.6\linewidth]{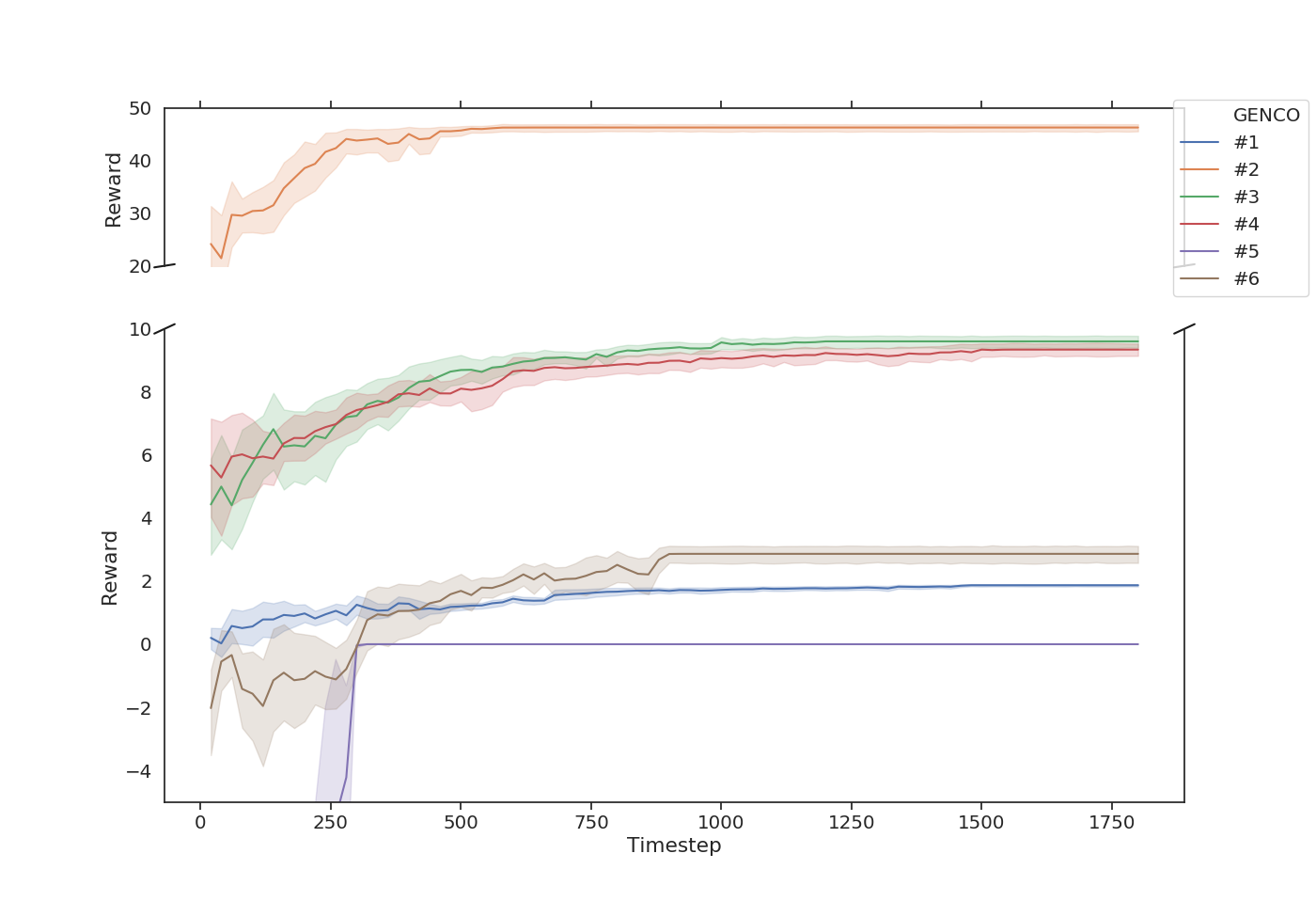}
    \caption{Learnt episodic rewards averaged across 10 random seeds with each of the GENCOs as a learner using NFQ-2 model and B-2 rival bidding strategy}
    \label{fig:rew_nfq2_b2}
\end{figure}

We consider each of the GENCOs as a learner and the bidding strategies of remaining 5 competing GENCOs are determined by B-2. This section presents the improvements achieved by each of the implementations of \textit{NFQ-TP} over the baseline true cost bidding.\\\\
\textbf{Performance of NFQ-1 variant}\\
Figure \ref{fig:rew_nfq1_b2} and Table \ref{tab:rew_nfq1_b2} show the average and standard deviation of episodic rewards across 10 random runs of the \textit{NFQ-TP} algorithm with each of the GENCOs as a learning agent and using NFQ-1 network implementation and B-2 rival bidding.
\\
\begin{table}[h]
    \caption{Mean $\mu$ and standard deviation $\sigma$ over 10 random runs using NFQ-1 model and B-2 rival bidding strategy}
    \centering
    \begin{tabular}{ccccccc}
    \toprule
        \# & 1 & 2 & 3 & 4 & 5 & 6\\
    \midrule
        $\mu$ & 1.22 & 25.99 & 6.01 & 6.53 & 0.0 & 0.0 \\
        $\sigma$ & 0.19 & 4.57 & 0.96 &  0.71 & 0.0 & 0.11\\
    \bottomrule
    \end{tabular}
    \label{tab:rew_nfq1_b2}
\end{table}\\
\textbf{Performance of NFQ-2 variant}\\
Figure \ref{fig:rew_nfq2_b2} and Table \ref{tab:rew_nfq2_b2} show the average and standard deviation of episodic rewards across 10 random runs of the \textit{NFQ-TP} algorithm with each of the GENCOs as a learning agent and using NFQ-2 network implementation and B-2 rival bidding.
\\
\begin{table}[h]
    \caption{Mean $\mu$ and standard deviation $\sigma$ over 10 random runs using NFQ-2 model and B-2 rival bidding strategy}
    \centering
    \begin{tabular}{ccccccc}
    \toprule
        \# & 1 & 2 & 3 & 4 & 5 & 6\\
    \midrule
        $\mu$ & 1.87 & 46.31 & 9.60 & 9.34 & 0.0 & 2.86 \\
        $\sigma$ & 0.06 & 1.13 & 0.31 & 0.33 & 0.0 & 0.44 \\
    \bottomrule
    \end{tabular}
    \label{tab:rew_nfq2_b2}
\end{table}
\subsubsection{\textbf{Comparison of NFQ-1 and NFQ-2.}}
On examining the mean episodic rewards for the two bidding strategies, we see that both the implementations are able to perform at least as well as the baseline bidding. However, we achieved far more superior results while using NFQ-2, even by using a smaller network. This is because of the way the network is structured. NFQ-1 maps a state-action pair to Q-value; hence, any update to the pair could affect the Q-values for other actions as well because of spatial locality. This also explains why the variance in episodic rewards is high even towards convergence while using NFQ-1 implementation.
\subsubsection{\textbf{Comparison of episodic rewards for B1 v/s B2.}} The mean episodic rewards achieved for the GENCOs are comparatively higher when the rival bidding strategy is B-2. This is because B-1 lets competitors follow an adaptive strategy based on the reactive power market requirement. Hence, it is more difficult for the learner to beat the rivals by a large margin in such a scenario. 
\subsubsection{\textbf{Observation and inference.}} The performance improvement achieved by \textit{NFQ-TP} algorithm using NFQ-2 network implementation is analysed. The variance in episodic rewards is high at the start of learning, meaning that the network is learning and exploring to find better bids. After around 1000 episodes, we see the variance reducing, and the episodic rewards converge, indicating that the algorithm has learnt the optimal bidding strategy. The episodic reward tables indicate that the approach is able to achieve significant improvement for intermediate and lower cost GENCOs while performing at least as well as baseline for the high-cost GENCOs. The use of multiple rival bidding strategies further shows that the method is suitable in different market scenarios. Therefore, \textit{NFQ-TP} provides an effective way for generation companies to improve their bidding strategies. 

\subsection{Impact of cost category of GENCO on learning benefits}
Learning benefits of GENCOs in each of the cost categories are also analysed. For ease of comparison, the analysis has been performed on NFQ-2 implementation for B-1 bidding strategy.
\subsubsection{Case 1: Low Cost GENCO as learner: }
\begin{figure}[h]
    \centering
    \includegraphics[width=0.6\linewidth]{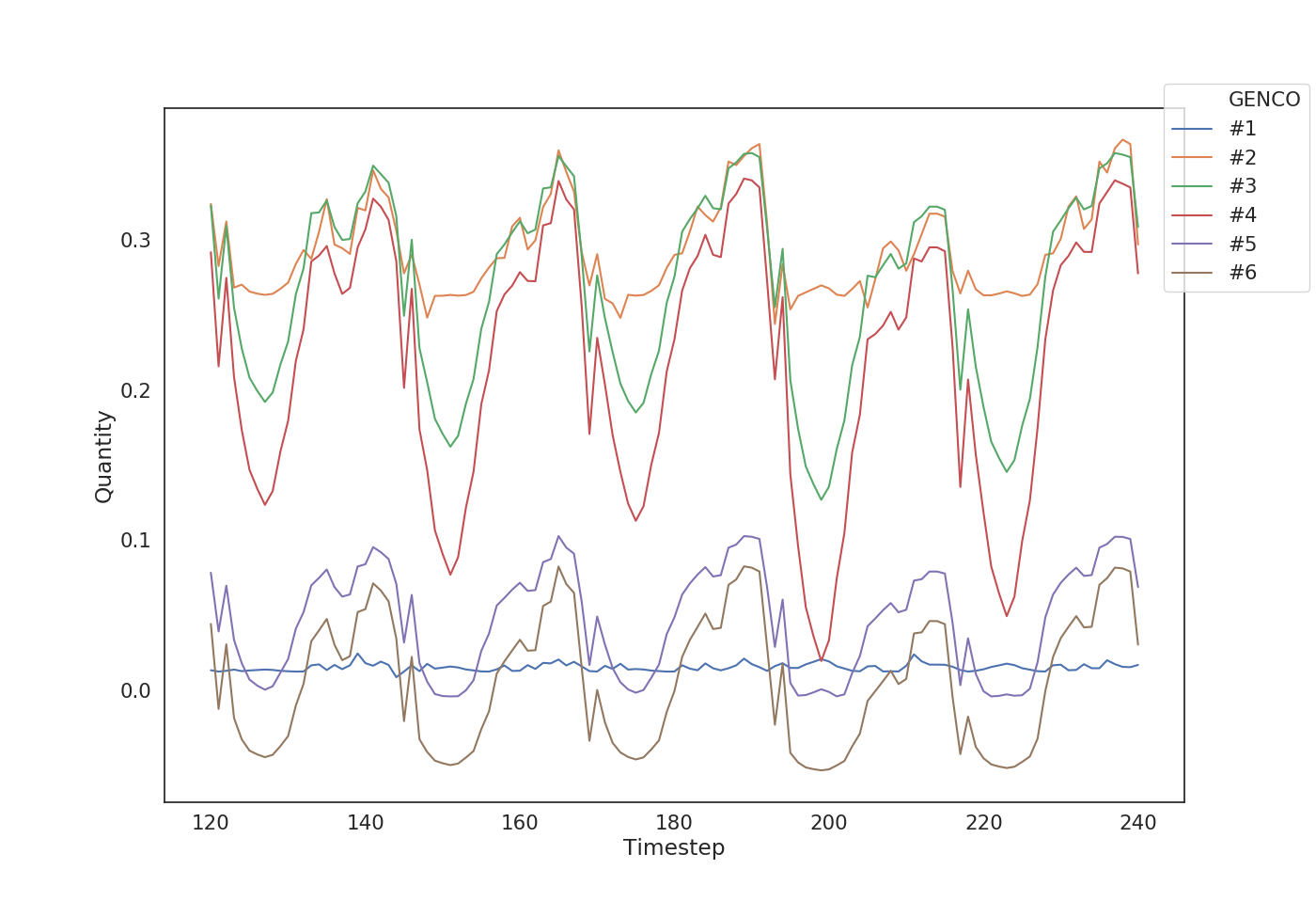}
    \caption{Quantity generated for all GENCOs across a 120 timestep window of the optimal bidding trajectory with GENCO-2 as learner}
    \label{fig:low_cost_win_qty}
\end{figure}
The generation curve for a 120 timestep window of the optimal bidding trajectory for GENCO-2 (Figure \ref{fig:low_cost_win_qty}) shows that low cost GENCO-2 not only peaks in high demand intervals by increasing its bidding cost coefficients but it also adjusts the bidding cost accordingly when the demand is lower and therefore, its generation does not dip as much in low demand intervals. The episodic reward plots further show that that \textit{NFQ-TP} is able to achieve significant improvements in episodic rewards for such GENCOs.
\subsubsection{Case 2: Intermediate Cost GENCO as learner: }
\begin{figure}[h]
    \centering
    \includegraphics[width=0.6\linewidth]{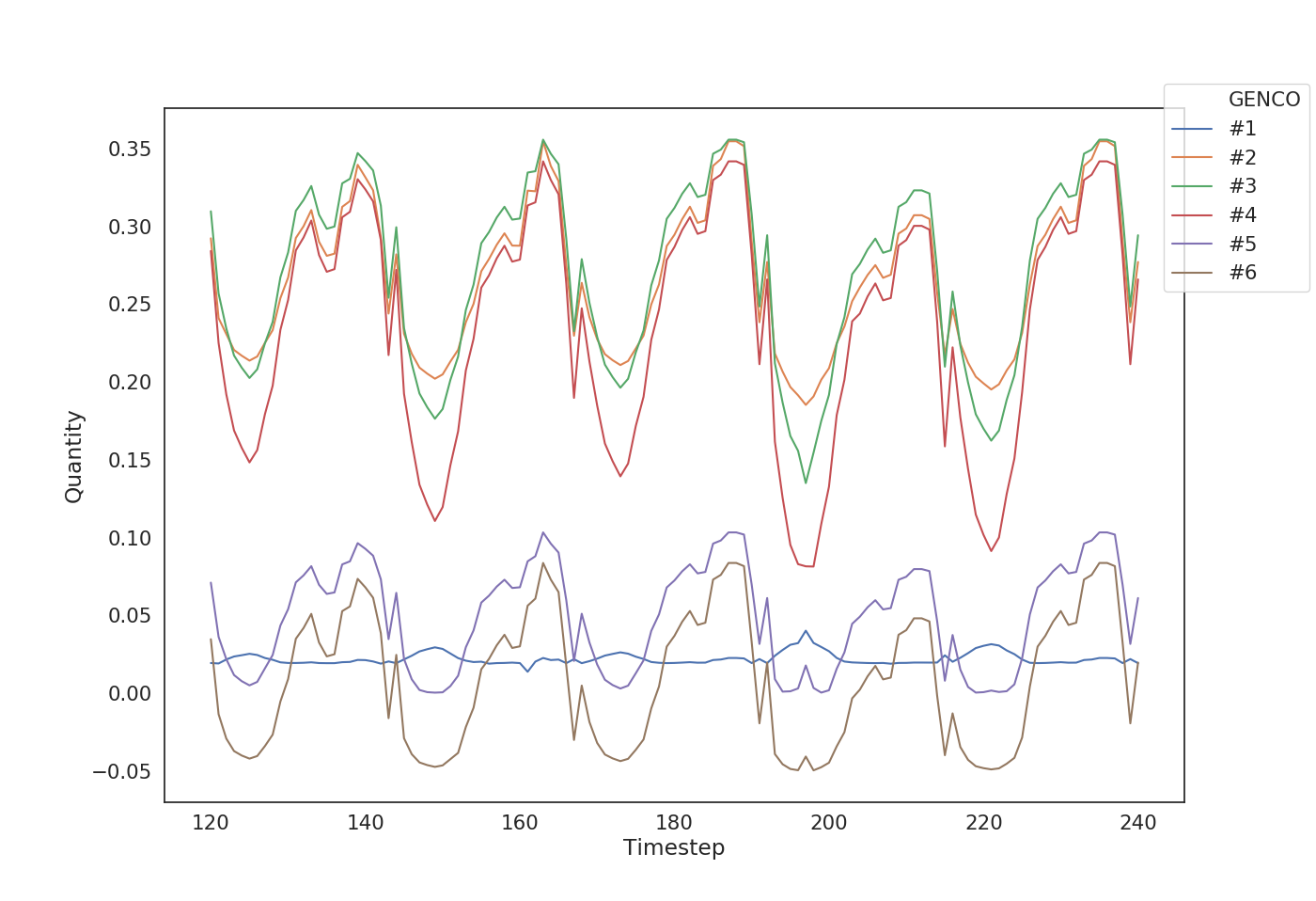}
    \caption{Quantity generated for all GENCOs across a 120 timestep window of the optimal bidding trajectory with GENCO-1 as learner}
    \label{fig:int_cost_win_qty}
\end{figure}
The generation curve for a 120 timestep window of the optimal bidding trajectory for GENCO-1 (Figure \ref{fig:int_cost_win_qty}) reveals that the learning is indeed useful especially in the times when demand is lower as the dips in generation quantity for Learner-1 are not as much as its competitors when it is learning and predicting the market demand. In such scenarios, the learner can exploit its knowledge to bid in such a way that the drop in quantity generated because of low demand is not as much as that of others. The episodic reward plots further show that that \textit{NFQ-TP} is able to achieve good improvements in episodic rewards for such GENCOs when the rivals are not as market aware (or optimal) and are able to achieve baseline bidding rewards in case of strong market-aware competitors.
\subsubsection{Case 3: High Cost GENCO as learner: }
\begin{figure}[h]
    \centering
    \includegraphics[width=0.6\linewidth]{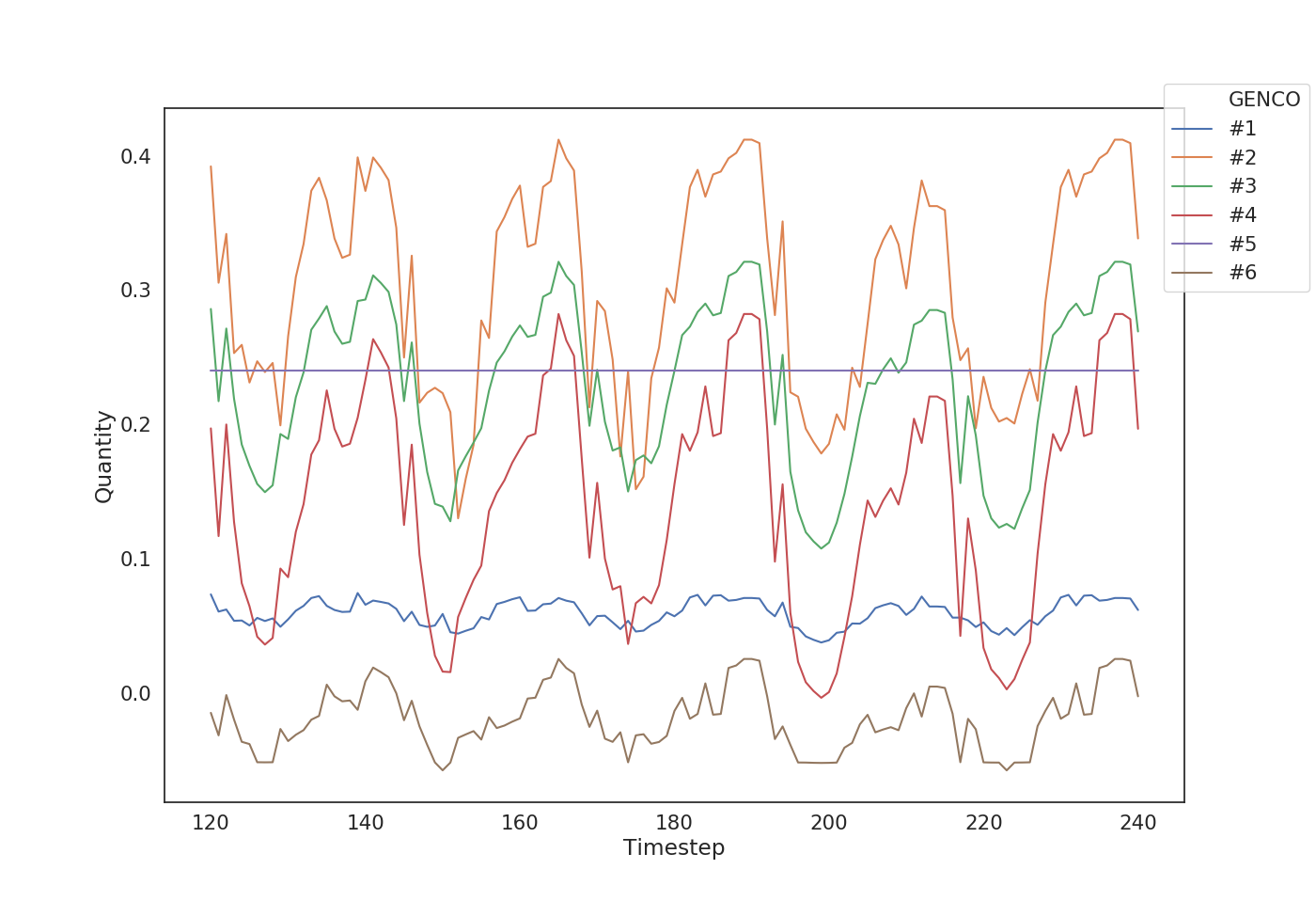}
    \caption{Quantity generated for all GENCOs across a 120 timestep window of the optimal bidding trajectory with GENCO-5 as learner}
    \label{fig:high_cost_win_qty}
\end{figure}
The generation curve for a 120 timestep window of the optimal bidding trajectory for GENCO-5 (Figure \ref{fig:high_cost_win_qty}) shows that high cost GENCOs try to maximize their profits by increasing the quantity they can generate. A comparison of quantity generation curve with the previous curves with other GENCOs as learners shows that the generation changed from about 0.025 to above 0.2 for GENCO-5 (which is around 10 times improvement). It is observed that, in general, the higher the actual costs of the GENCOs, closer the bid values are to the actual costs of generation so as to achieve higher profits. This is justified because higher the cost a GENCO bids, more is the chance that the ISO might assign higher quantity to competing GENCOs. 

\section{Conclusion and Future Work}
The three-stage reactive power market model is designed to encourage GENCOs to bid closer to their true cost and prevent any single agent from gaining market power. The invisibility of rival bids and difficulty in estimating bid magnification from the price signals received by ISO presents challenges in directly estimating the optimal bid value. The environment that we propose takes features directly from the market observations such as bids and the corresponding rewards and learns some using experience, such as the demand. This work showed techniques to successfully compute optimal bidding strategy for GENCOs in different scenarios: imperfect competition, when the other agents have a time-dependent strategy vs. perfect competition, when they follow a fairly regular pattern of bidding their true cost. 

It was shown that using a Q-value estimation technique like Neural Fitted Q Iteration coupled with our own experience and market observations, GENCOs can bid a value which is a magnification of their actual costs and in some scenarios, can still make more profit than telling the truth. This is because our learners are more aware about the market behaviour and hence, adapt their costs according to the market requirements. Because learning becomes more difficult when bid values for other GENCOs are hidden, this work shows why the restriction on visibility of bid values prevents GENCOs from exercising market power. 

The work presented in the paper considered rivals of the learning agent to follow stochastic optimisation bidding strategy. This can be further improved by considering similar NFQ-TP learning algorithms for rivals as well to extend the work to a multi-agent learning environment after a few relaxations. Another interesting experiment is to analyse how market behaviour and optimal bidding strategy are affected if market mechanisms are separate for low-cost GENCOs (i.e., renewable energy sources) and high-cost GENCOs (i.e., conventional or non-renewable sources). This would help us determine if separate markets led to lower total cost for ISO as the low-cost GENCOs are free to bid higher in the common market and can take more advantage of the ISO. 

% \section{Note:}

% Two elements of the ``acmart'' document class provide powerful
% taxonomic tools for you to help readers find your work in an online
% search.

% The ACM Computing Classification System ---
% \url{https://www.acm.org/publications/class-2012} --- is a set of
% classifiers and concepts that describe the computing
% discipline. Authors can select entries from this classification
% system, via \url{https://dl.acm.org/ccs/ccs.cfm}, and generate the
% commands to be included in the \LaTeX\ source.

% User-defined keywords are a comma-separated list of words and phrases
% of the authors' choosing, providing a more flexible way of describing
% the research being presented.

% CCS concepts and user-defined keywords are required for for all
% articles over two pages in length, and are optional for one- and
% two-page articles (or abstracts).

%%
%% The next two lines define the bibliography style to be used, and
%% the bibliography file.
\bibliographystyle{ACM-Reference-Format}
\bibliography{sample-acmsmall}

%%% -*-BibTeX-*-
%%% Do NOT edit. File created by BibTeX with style
%%% ACM-Reference-Format-Journals [18-Jan-2012].

\begin{thebibliography}{24}

%%% ====================================================================
%%% NOTE TO THE USER: you can override these defaults by providing
%%% customized versions of any of these macros before the \bibliography
%%% command.  Each of them MUST provide its own final punctuation,
%%% except for \shownote{}, \showDOI{}, and \showURL{}.  The latter two
%%% do not use final punctuation, in order to avoid confusing it with
%%% the Web address.
%%%
%%% To suppress output of a particular field, define its macro to expand
%%% to an empty string, or better, \unskip, like this:
%%%
%%% \newcommand{\showDOI}[1]{\unskip}   % LaTeX syntax
%%%
%%% \def \showDOI #1{\unskip}           % plain TeX syntax
%%%
%%% ====================================================================

\ifx \showCODEN    \undefined \def \showCODEN     #1{\unskip}     \fi
\ifx \showDOI      \undefined \def \showDOI       #1{#1}\fi
\ifx \showISBNx    \undefined \def \showISBNx     #1{\unskip}     \fi
\ifx \showISBNxiii \undefined \def \showISBNxiii  #1{\unskip}     \fi
\ifx \showISSN     \undefined \def \showISSN      #1{\unskip}     \fi
\ifx \showLCCN     \undefined \def \showLCCN      #1{\unskip}     \fi
\ifx \shownote     \undefined \def \shownote      #1{#1}          \fi
\ifx \showarticletitle \undefined \def \showarticletitle #1{#1}   \fi
\ifx \showURL      \undefined \def \showURL       {\relax}        \fi
% The following commands are used for tagged output and should be
% invisible to TeX
\providecommand\bibfield[2]{#2}
\providecommand\bibinfo[2]{#2}
\providecommand\natexlab[1]{#1}
\providecommand\showeprint[2][]{arXiv:#2}

\bibitem[\protect\citeauthoryear{Badri and Rashidinejad}{Badri and
  Rashidinejad}{2013}]%
        {badri2013security}
\bibfield{author}{\bibinfo{person}{A Badri} {and} \bibinfo{person}{M
  Rashidinejad}.} \bibinfo{year}{2013}\natexlab{}.
\newblock \showarticletitle{Security constrained optimal bidding strategy of
  GenCos in day ahead oligopolistic power markets: a Cournot-based model}.
\newblock \bibinfo{journal}{\emph{Electrical Engineering}}
  \bibinfo{volume}{95}, \bibinfo{number}{2} (\bibinfo{year}{2013}),
  \bibinfo{pages}{63--72}.
\newblock


\bibitem[\protect\citeauthoryear{Cao, Hu, Xu, Dragi{\v{c}}evi{\'c}, Huang, Liu,
  Chen, and Blaabjerg}{Cao et~al\mbox{.}}{2020}]%
        {cao2020bidding}
\bibfield{author}{\bibinfo{person}{Di Cao}, \bibinfo{person}{Weihao Hu},
  \bibinfo{person}{Xiao Xu}, \bibinfo{person}{Tomislav Dragi{\v{c}}evi{\'c}},
  \bibinfo{person}{Qi Huang}, \bibinfo{person}{Zhou Liu}, \bibinfo{person}{Zhe
  Chen}, {and} \bibinfo{person}{Frede Blaabjerg}.}
  \bibinfo{year}{2020}\natexlab{}.
\newblock \showarticletitle{Bidding strategy for trading wind energy and
  purchasing reserve of wind power producer--A DRL based approach}.
\newblock \bibinfo{journal}{\emph{International Journal of Electrical Power \&
  Energy Systems}}  \bibinfo{volume}{117} (\bibinfo{year}{2020}),
  \bibinfo{pages}{105648}.
\newblock


\bibitem[\protect\citeauthoryear{Chen, Paschalidis, Caramanis, and
  Andrianesis}{Chen et~al\mbox{.}}{2019}]%
        {past_bids}
\bibfield{author}{\bibinfo{person}{Ruidi Chen}, \bibinfo{person}{Ioannis~Ch
  Paschalidis}, \bibinfo{person}{Michael~C Caramanis}, {and}
  \bibinfo{person}{Panagiotis Andrianesis}.} \bibinfo{year}{2019}\natexlab{}.
\newblock \showarticletitle{Learning from past bids to participate
  strategically in day-ahead electricity markets}.
\newblock \bibinfo{journal}{\emph{IEEE Transactions on Smart Grid}}
  \bibinfo{volume}{10}, \bibinfo{number}{5} (\bibinfo{year}{2019}),
  \bibinfo{pages}{5794--5806}.
\newblock


\bibitem[\protect\citeauthoryear{Chitkara, Zhong, and Bhattacharya}{Chitkara
  et~al\mbox{.}}{2009}]%
        {oligopoly}
\bibfield{author}{\bibinfo{person}{Puneet Chitkara}, \bibinfo{person}{Jin
  Zhong}, {and} \bibinfo{person}{Kankar Bhattacharya}.}
  \bibinfo{year}{2009}\natexlab{}.
\newblock \showarticletitle{Oligopolistic competition of gencos in reactive
  power ancillary service provisions}.
\newblock \bibinfo{journal}{\emph{IEEE Transactions on Power Systems}}
  \bibinfo{volume}{24}, \bibinfo{number}{3} (\bibinfo{year}{2009}),
  \bibinfo{pages}{1256--1265}.
\newblock


\bibitem[\protect\citeauthoryear{Commission et~al\mbox{.}}{Commission
  et~al\mbox{.}}{1996}]%
        {us1996promoting}
\bibfield{author}{\bibinfo{person}{US~Federal Energy~Regulatory Commission}
  {et~al\mbox{.}}} \bibinfo{year}{1996}\natexlab{}.
\newblock \showarticletitle{Promoting wholesale competition through open access
  non-discriminatory transmission services by public utilities; recovery of
  stranded costs by public utilities and transmitting utilities}.
\newblock \bibinfo{journal}{\emph{Order}}  \bibinfo{volume}{888}
  (\bibinfo{year}{1996}), \bibinfo{pages}{24}.
\newblock


\bibitem[\protect\citeauthoryear{Ernst, Geurts, and Wehenkel}{Ernst
  et~al\mbox{.}}{2005}]%
        {ernst2005tree}
\bibfield{author}{\bibinfo{person}{Damien Ernst}, \bibinfo{person}{Pierre
  Geurts}, {and} \bibinfo{person}{Louis Wehenkel}.}
  \bibinfo{year}{2005}\natexlab{}.
\newblock \showarticletitle{Tree-based batch mode reinforcement learning}.
\newblock \bibinfo{journal}{\emph{Journal of Machine Learning Research}}
  \bibinfo{volume}{6}, \bibinfo{number}{Apr} (\bibinfo{year}{2005}),
  \bibinfo{pages}{503--556}.
\newblock


\bibitem[\protect\citeauthoryear{Gallego, Duarte, and Delgadillo}{Gallego
  et~al\mbox{.}}{2008}]%
        {colombian}
\bibfield{author}{\bibinfo{person}{L Gallego}, \bibinfo{person}{O Duarte},
  {and} \bibinfo{person}{A Delgadillo}.} \bibinfo{year}{2008}\natexlab{}.
\newblock \showarticletitle{Strategic bidding in Colombian electricity market
  using a multi-agent learning approach}. In \bibinfo{booktitle}{\emph{2008
  IEEE/PES Transmission and Distribution Conference and Exposition: Latin
  America}}. IEEE, \bibinfo{pages}{1--7}.
\newblock


\bibitem[\protect\citeauthoryear{{Jay} and {SWARUP}}{{Jay} and
  {SWARUP}}{2020}]%
        {9179006}
\bibfield{author}{\bibinfo{person}{D. {Jay}} {and} \bibinfo{person}{S.~K.
  {SWARUP}}.} \bibinfo{year}{2020}\natexlab{}.
\newblock \showarticletitle{Game Theoretical Approach to Novel Reactive Power
  Ancillary Service Market Mechanism}.
\newblock \bibinfo{journal}{\emph{IEEE Transactions on Power Systems}}
  (\bibinfo{year}{2020}), \bibinfo{pages}{1--1}.
\newblock


\bibitem[\protect\citeauthoryear{{Jiang}, {Hou}, {Lin}, {Wen}, {Li}, {He},
  {Ji}, {Lin}, {Ding}, and {Yang}}{{Jiang} et~al\mbox{.}}{2019}]%
        {8723337}
\bibfield{author}{\bibinfo{person}{Y. {Jiang}}, \bibinfo{person}{J. {Hou}},
  \bibinfo{person}{Z. {Lin}}, \bibinfo{person}{F. {Wen}}, \bibinfo{person}{J.
  {Li}}, \bibinfo{person}{C. {He}}, \bibinfo{person}{C. {Ji}},
  \bibinfo{person}{Z. {Lin}}, \bibinfo{person}{Y. {Ding}}, {and}
  \bibinfo{person}{L. {Yang}}.} \bibinfo{year}{2019}\natexlab{}.
\newblock \showarticletitle{Optimal Bidding Strategy for a Power Producer Under
  Monthly Pre-Listing Balancing Mechanism in Actual Sequential Energy
  Dual-Market in China}.
\newblock \bibinfo{journal}{\emph{IEEE Access}}  \bibinfo{volume}{7}
  (\bibinfo{year}{2019}), \bibinfo{pages}{70986--70998}.
\newblock


\bibitem[\protect\citeauthoryear{Leyffer and Munson}{Leyffer and
  Munson}{2005}]%
        {mul_leader}
\bibfield{author}{\bibinfo{person}{Sven Leyffer} {and} \bibinfo{person}{Todd
  Munson}.} \bibinfo{year}{2005}\natexlab{}.
\newblock \showarticletitle{Solving multi-leader-follower games}.
\newblock \bibinfo{journal}{\emph{Preprint ANL/MCS-P1243-0405}}
  \bibinfo{volume}{4}, \bibinfo{number}{04} (\bibinfo{year}{2005}).
\newblock


\bibitem[\protect\citeauthoryear{Mathur, Arya, and Dubey}{Mathur
  et~al\mbox{.}}{2017}]%
        {mathur2017optimal}
\bibfield{author}{\bibinfo{person}{Somendra~PS Mathur}, \bibinfo{person}{Anoop
  Arya}, {and} \bibinfo{person}{Manisha Dubey}.}
  \bibinfo{year}{2017}\natexlab{}.
\newblock \showarticletitle{Optimal bidding strategy for price takers and
  customers in a competitive electricity market}.
\newblock \bibinfo{journal}{\emph{Cogent Engineering}} \bibinfo{volume}{4},
  \bibinfo{number}{1} (\bibinfo{year}{2017}), \bibinfo{pages}{1358545}.
\newblock


\bibitem[\protect\citeauthoryear{Nanduri and Das}{Nanduri and Das}{2007}]%
        {nanduri2007reinforcement}
\bibfield{author}{\bibinfo{person}{Vishnuteja Nanduri} {and}
  \bibinfo{person}{Tapas~K Das}.} \bibinfo{year}{2007}\natexlab{}.
\newblock \showarticletitle{A reinforcement learning model to assess market
  power under auction-based energy pricing}.
\newblock \bibinfo{journal}{\emph{IEEE transactions on Power Systems}}
  \bibinfo{volume}{22}, \bibinfo{number}{1} (\bibinfo{year}{2007}),
  \bibinfo{pages}{85--95}.
\newblock


\bibitem[\protect\citeauthoryear{Rahimiyan and Mashhadi}{Rahimiyan and
  Mashhadi}{2010}]%
        {rahimiyan2010adaptive}
\bibfield{author}{\bibinfo{person}{Morteza Rahimiyan} {and}
  \bibinfo{person}{Habib~Rajabi Mashhadi}.} \bibinfo{year}{2010}\natexlab{}.
\newblock \showarticletitle{An Adaptive $ Q $-Learning Algorithm Developed for
  Agent-Based Computational Modeling of Electricity Market}.
\newblock \bibinfo{journal}{\emph{IEEE Transactions on Systems, Man, and
  Cybernetics, Part C (Applications and Reviews)}} \bibinfo{volume}{40},
  \bibinfo{number}{5} (\bibinfo{year}{2010}), \bibinfo{pages}{547--556}.
\newblock


\bibitem[\protect\citeauthoryear{Rayati, Goodarzi, and Ranjbar}{Rayati
  et~al\mbox{.}}{2019}]%
        {rayati2019optimal}
\bibfield{author}{\bibinfo{person}{Mohammad Rayati}, \bibinfo{person}{Hamed
  Goodarzi}, {and} \bibinfo{person}{AliMohammad Ranjbar}.}
  \bibinfo{year}{2019}\natexlab{}.
\newblock \showarticletitle{Optimal bidding strategy of coordinated wind power
  and gas turbine units in real-time market using conditional value at risk}.
\newblock \bibinfo{journal}{\emph{International Transactions on Electrical
  Energy Systems}} \bibinfo{volume}{29}, \bibinfo{number}{1}
  (\bibinfo{year}{2019}), \bibinfo{pages}{e2645}.
\newblock


\bibitem[\protect\citeauthoryear{Riedmiller}{Riedmiller}{2005}]%
        {riedmiller2005neural}
\bibfield{author}{\bibinfo{person}{Martin Riedmiller}.}
  \bibinfo{year}{2005}\natexlab{}.
\newblock \showarticletitle{Neural fitted Q iteration--first experiences with a
  data efficient neural reinforcement learning method}. In
  \bibinfo{booktitle}{\emph{European Conference on Machine Learning}}.
  Springer, \bibinfo{pages}{317--328}.
\newblock


\bibitem[\protect\citeauthoryear{Riedmiller and Braun}{Riedmiller and
  Braun}{1993}]%
        {riedmiller1993direct}
\bibfield{author}{\bibinfo{person}{Martin Riedmiller} {and}
  \bibinfo{person}{Heinrich Braun}.} \bibinfo{year}{1993}\natexlab{}.
\newblock \showarticletitle{A direct adaptive method for faster backpropagation
  learning: The RPROP algorithm}. In \bibinfo{booktitle}{\emph{IEEE
  international conference on neural networks}}. IEEE,
  \bibinfo{pages}{586--591}.
\newblock


\bibitem[\protect\citeauthoryear{Schaul, Quan, Antonoglou, and Silver}{Schaul
  et~al\mbox{.}}{2015}]%
        {schaul2015prioritized}
\bibfield{author}{\bibinfo{person}{Tom Schaul}, \bibinfo{person}{John Quan},
  \bibinfo{person}{Ioannis Antonoglou}, {and} \bibinfo{person}{David Silver}.}
  \bibinfo{year}{2015}\natexlab{}.
\newblock \showarticletitle{Prioritized experience replay}.
\newblock \bibinfo{journal}{\emph{arXiv preprint arXiv:1511.05952}}
  (\bibinfo{year}{2015}).
\newblock


\bibitem[\protect\citeauthoryear{Singh and Fozdar}{Singh and Fozdar}{2019}]%
        {singh2019optimal}
\bibfield{author}{\bibinfo{person}{Satyendra Singh} {and}
  \bibinfo{person}{Manoj Fozdar}.} \bibinfo{year}{2019}\natexlab{}.
\newblock \showarticletitle{Optimal bidding strategy with the inclusion of wind
  power supplier in an emerging power market}.
\newblock \bibinfo{journal}{\emph{IET Generation, Transmission \&
  Distribution}} \bibinfo{volume}{13}, \bibinfo{number}{10}
  (\bibinfo{year}{2019}), \bibinfo{pages}{1914--1922}.
\newblock


\bibitem[\protect\citeauthoryear{Subramanian, Bichpuriya, Achar, Bhat, Singh,
  Sarangan, and Natarajan}{Subramanian et~al\mbox{.}}{2019}]%
        {learn}
\bibfield{author}{\bibinfo{person}{Easwar Subramanian}, \bibinfo{person}{Yogesh
  Bichpuriya}, \bibinfo{person}{Avinash Achar}, \bibinfo{person}{Sanjay Bhat},
  \bibinfo{person}{Abhay~Pratap Singh}, \bibinfo{person}{Venkatesh Sarangan},
  {and} \bibinfo{person}{Akshaya Natarajan}.} \bibinfo{year}{2019}\natexlab{}.
\newblock \showarticletitle{lEarn: A Reinforcement Learning Based Bidding
  Strategy for Generators in Single sided Energy Markets}. In
  \bibinfo{booktitle}{\emph{Proceedings of the Tenth ACM International
  Conference on Future Energy Systems}}. \bibinfo{pages}{121--127}.
\newblock


\bibitem[\protect\citeauthoryear{Watkins and Dayan}{Watkins and Dayan}{1992}]%
        {watkins1992q}
\bibfield{author}{\bibinfo{person}{Christopher~JCH Watkins} {and}
  \bibinfo{person}{Peter Dayan}.} \bibinfo{year}{1992}\natexlab{}.
\newblock \showarticletitle{Q-learning}.
\newblock \bibinfo{journal}{\emph{Machine learning}} \bibinfo{volume}{8},
  \bibinfo{number}{3-4} (\bibinfo{year}{1992}), \bibinfo{pages}{279--292}.
\newblock


\bibitem[\protect\citeauthoryear{Wen and David}{Wen and David}{2001}]%
        {wen2001genetic}
\bibfield{author}{\bibinfo{person}{Fushuan Wen} {and} \bibinfo{person}{A~Kumar
  David}.} \bibinfo{year}{2001}\natexlab{}.
\newblock \showarticletitle{A genetic algorithm based method for bidding
  strategy coordination in energy and spinning reserve markets}.
\newblock \bibinfo{journal}{\emph{Artificial Intelligence in Engineering}}
  \bibinfo{volume}{15}, \bibinfo{number}{1} (\bibinfo{year}{2001}),
  \bibinfo{pages}{71--79}.
\newblock


\bibitem[\protect\citeauthoryear{Wozabal and Rameseder}{Wozabal and
  Rameseder}{2020}]%
        {wozabal2020optimal}
\bibfield{author}{\bibinfo{person}{David Wozabal} {and}
  \bibinfo{person}{Gunther Rameseder}.} \bibinfo{year}{2020}\natexlab{}.
\newblock \showarticletitle{Optimal bidding of a virtual power plant on the
  Spanish day-ahead and intraday market for electricity}.
\newblock \bibinfo{journal}{\emph{European Journal of Operational Research}}
  \bibinfo{volume}{280}, \bibinfo{number}{2} (\bibinfo{year}{2020}),
  \bibinfo{pages}{639--655}.
\newblock


\bibitem[\protect\citeauthoryear{Ye, Qiu, Sun, Papadaskalopoulos, and
  Strbac}{Ye et~al\mbox{.}}{2019}]%
        {ye2019deep}
\bibfield{author}{\bibinfo{person}{Yujian Ye}, \bibinfo{person}{Dawei Qiu},
  \bibinfo{person}{Mingyang Sun}, \bibinfo{person}{Dimitrios
  Papadaskalopoulos}, {and} \bibinfo{person}{Goran Strbac}.}
  \bibinfo{year}{2019}\natexlab{}.
\newblock \showarticletitle{Deep reinforcement learning for strategic bidding
  in electricity markets}.
\newblock \bibinfo{journal}{\emph{IEEE Transactions on Smart Grid}}
  \bibinfo{volume}{11}, \bibinfo{number}{2} (\bibinfo{year}{2019}),
  \bibinfo{pages}{1343--1355}.
\newblock


\bibitem[\protect\citeauthoryear{Zimmerman, Murillo-S{\'a}nchez, and
  Thomas}{Zimmerman et~al\mbox{.}}{2010}]%
        {zimmerman2010matpower}
\bibfield{author}{\bibinfo{person}{Ray~Daniel Zimmerman},
  \bibinfo{person}{Carlos~Edmundo Murillo-S{\'a}nchez}, {and}
  \bibinfo{person}{Robert~John Thomas}.} \bibinfo{year}{2010}\natexlab{}.
\newblock \showarticletitle{MATPOWER: Steady-state operations, planning, and
  analysis tools for power systems research and education}.
\newblock \bibinfo{journal}{\emph{IEEE Transactions on power systems}}
  \bibinfo{volume}{26}, \bibinfo{number}{1} (\bibinfo{year}{2010}),
  \bibinfo{pages}{12--19}.
\newblock


\end{thebibliography}

%%
%% If your work has an appendix, this is the place to put it.
\appendix
\section{Hyperparameter Tuning}
\label{sec:hyp_tune}
\begin{figure}
     \centering
     \begin{tabular}{cc}
\begin{subfigure}[h]{0.5\textwidth}
   \centering
    \includegraphics[width=0.8\linewidth]{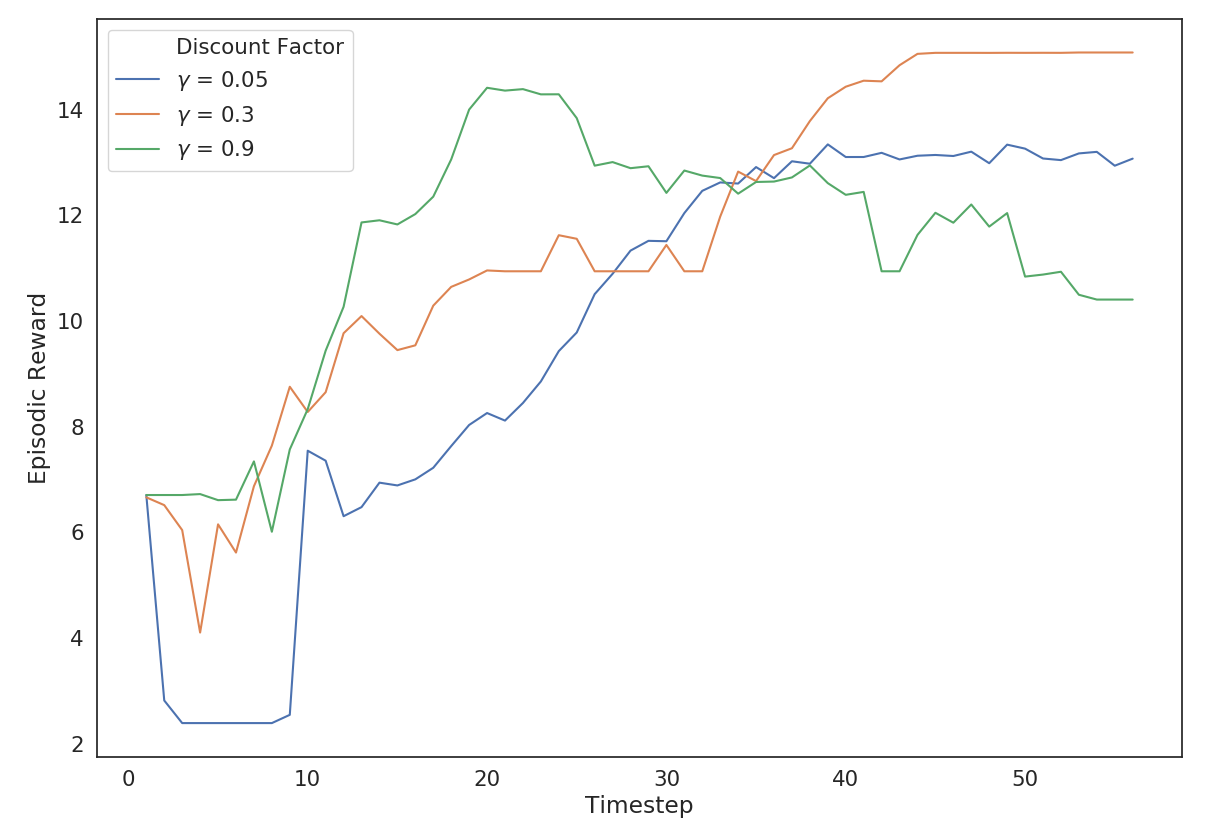}
    \caption{Effect of discount factor on episodic rewards}
    \label{fig:df_var}
\end{subfigure}
\hfill
\begin{subfigure}[h]{0.5\textwidth}
    \centering
    \includegraphics[width=0.8\linewidth]{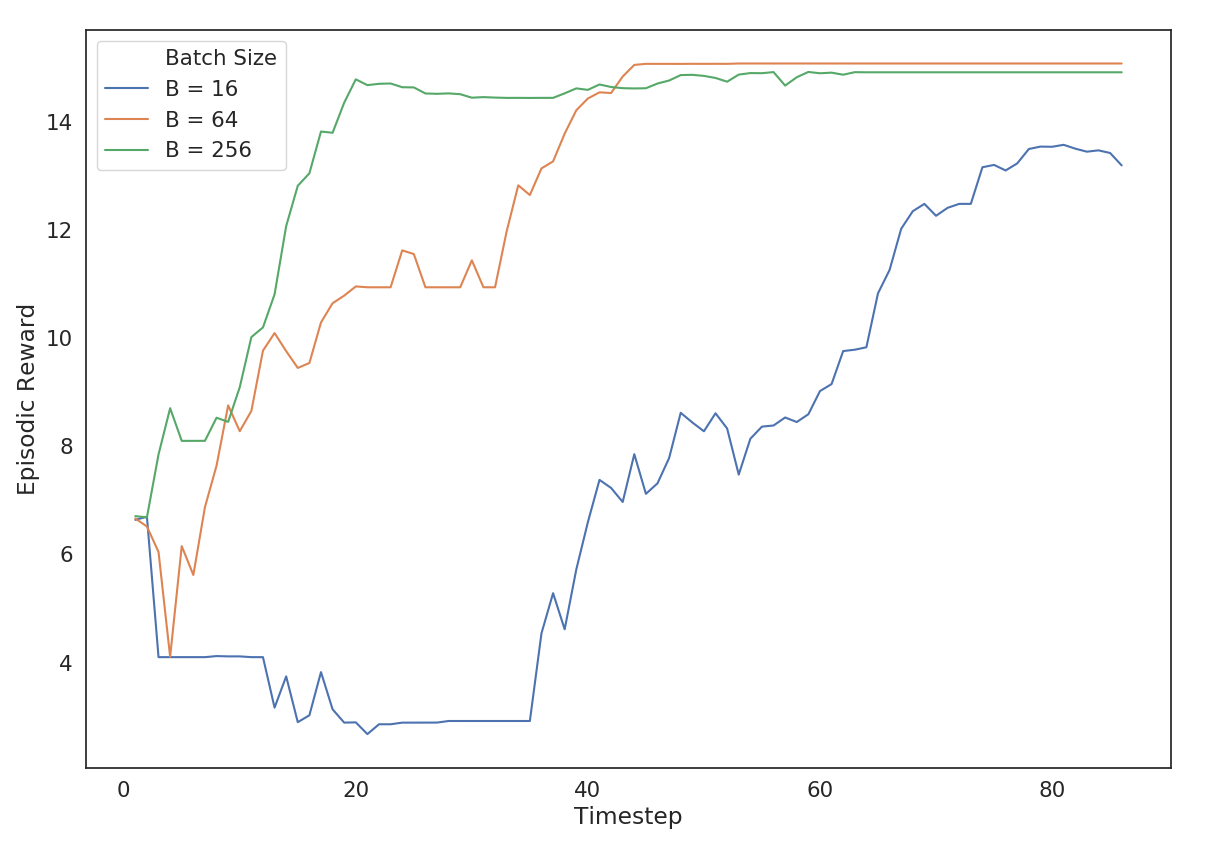}
    \caption{Effect of mini-batch size on episodic rewards}
    \label{fig:bs_var}
\end{subfigure}
       \end{tabular}
\begin{subfigure}[h]{0.5\textwidth}
    \centering
    \includegraphics[width=0.8\linewidth]{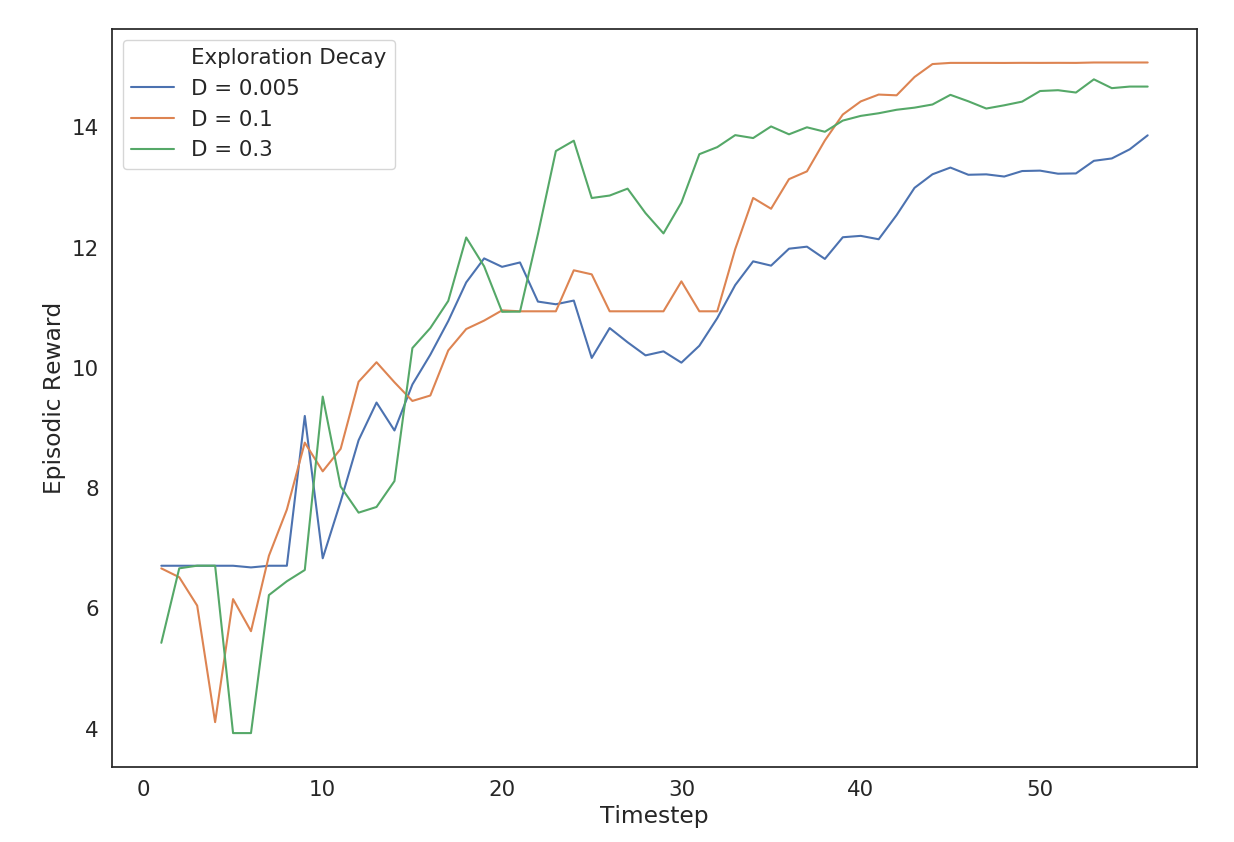}
    \caption{Effect of exploration decay on episodic rewards}
    \label{fig:ed_var}
\end{subfigure}
\caption{Effect of hyper-parameters on episodic rewards achieved by the learning agent}
\end{figure}

This section aims to present the selection process behind the crucial hyper-parameters. Because of the large number of parameters involved in \textit{NFQ-TP} algorithm, it is difficult to perform a grid-search, hence, the parameters have been chosen by manual search. For comparison, all the experiments use NFQ-2 implementation with B-1 rival bidding strategy. We first look at how the discount factor $\gamma$ affects the episodic rewards for an arbitrarily chosen GENCO (say, 2) in Figure \ref{fig:df_var}. The second parameter that we will analyze is the size of mini-batch sampled from Prioritized Experience Replay in Figure \ref{fig:bs_var}. Finally, we examine the rate at which exploration factor $\epsilon$ should be decayed in Figure \ref{fig:ed_var}. 
\\\\
\textbf{Observation and inference}\\
A high value of $\gamma = 0.9$ leads to focusing more on future rewards and hence, the Q-values diverge. This is because when we use bootstrapping, our Q-value estimates are not very accurate and hence, focusing on them more leads to instability in learning. A low value of $\gamma = 0.05$ makes the learner myopic and hence, learns a less than optimal bidding strategy. A discount factor of 0.3 give best observed episodic rewards.

A high value of batch size (256) means slower learning and poor generalization leading to lesser episodic rewards. A low value of batch size (say, 16), on the other hand, could lead to overfitting on the batch and hence, convergence to local optima. A batch size of 64 give best observed episodic rewards.

A very high exploration decay rate leads to high exploitation and less exploration quickly and hence, the learner converges at a sub-optimal policy. A very low value of exploration decay factor could lead to excessive exploration and hence, slow learning. An exploration decay rate of 0.1 gives best observed episodic rewards.

\end{document}